\documentclass[acmsmall,screen]{acmart}

\AtBeginDocument{%
  }

\setcopyright{acmcopyright}
\copyrightyear{2023}
\acmYear{2023}
\acmDOI{XXXXXXX.XXXXXXX}

\acmJournal{TOIT}
\acmVolume{1}
\acmNumber{1}
\acmArticle{1}
\acmMonth{1}

\usepackage{multirow}
\usepackage{makecell}
\usepackage{colortbl}
\usepackage{pifont}
\usepackage{diagbox}
\usepackage{newtxmath}

\definecolor{gray}{HTML}{E0E0E0}
\newcommand{\xmark}{\ding{53}}

\newcommand{\rw}[1]{\textcolor{black}{#1}}

\begin{document}

\title{Open Set Dandelion Network for IoT Intrusion Detection}

\author{Jiashu Wu}
\email{wujiashu21@mails.ucas.ac.cn}
\orcid{0000-0002-1347-1974}
\author{Hao Dai}
\email{daihao19@mails.ucas.ac.cn}
\orcid{0000-0003-1018-2162}
\affiliation{%
  \institution{Shenzhen Institute of Advanced Technology, Chinese Academy of Sciences}
  \city{Shenzhen}
  \state{Guangdong}
  \country{China}
  \postcode{518055}
}
\affiliation{%
  \institution{University of Chinese Academy of Sciences}
  \city{Beijing}
  \state{Beijing}
  \country{China}
  \postcode{100049}
}

\author{Kenneth B. Kent}
\email{ken@unb.ca}
\affiliation{%
  \institution{University of New Brunswick}
  \city{Fredericton}
  \state{New Brunswick}
  \country{Canada}
  \postcode{E3B 5A3}
}

\author{Jerome Yen}
\email{jeromeyen@um.edu.mo}
\affiliation{%
  \institution{University of Macau}
  \city{Taipa}
  \state{Macau}
  \country{China}
  \postcode{999078}
}

\author{Chengzhong Xu}
\email{czxu@um.edu.mo}
\orcid{0000-0001-9480-0356}
\affiliation{%
  \institution{University of Macau}
  \city{Taipa}
  \state{Macau}
  \country{China}
  \postcode{999078}
}

\author{Yang Wang}
\authornote{Corresponding Author. Email: yang.wang1@siat.ac.cn}
\email{yang.wang1@siat.ac.cn}
\orcid{0000-0001-9438-6060}
\affiliation{%
  \institution{Shenzhen Institute of Advanced Technology, Chinese Academy of Sciences}
  \city{Shenzhen}
  \state{Guangdong}
  \country{China}
  \postcode{518055}
}

\renewcommand{\shortauthors}{Jiashu Wu et al.}

\begin{abstract}
  As Internet of Things devices become widely used in the real world, it is crucial to protect them from malicious intrusions. However, the data scarcity of IoT limits the applicability of traditional intrusion detection methods, \rw{which are highly data-dependent}. To address this, in this paper we propose the Open-Set Dandelion Network (OSDN) based on unsupervised heterogeneous domain adaptation in an open-set manner. The OSDN model performs intrusion knowledge transfer from the knowledge-rich source network intrusion domain to facilitate more accurate intrusion detection for the data-scarce target IoT intrusion domain. Under the open-set setting, it can also detect newly-emerged target domain intrusions that are not observed in the source domain. To achieve this, the OSDN model forms the source domain into a dandelion-like feature space in which \rw{each intrusion category is compactly grouped and different intrusion categories are separated, i.e., simultaneously emphasising inter-category separability and intra-category compactness. }The dandelion-based target membership mechanism then forms the target dandelion. Then, the dandelion angular separation mechanism achieves better inter-category separability, and the dandelion embedding alignment mechanism further aligns both dandelions in a finer manner. To promote intra-category compactness, the discriminating sampled dandelion mechanism is used. Assisted by the intrusion classifier trained using both known and generated unknown intrusion knowledge, a semantic dandelion correction mechanism emphasises easily-confused categories and guides better inter-category separability. Holistically, these mechanisms form the OSDN model that effectively performs intrusion knowledge transfer to benefit IoT intrusion detection. Comprehensive experiments on several intrusion datasets verify the effectiveness of the OSDN model, outperforming three state-of-the-art baseline methods by $16.9\%$. The contribution of each OSDN constituting component, the stability and the efficiency of the OSDN model \rw{are} also verified. 
\end{abstract}

\ccsdesc[500]{Computing methodologies~Machine learning}
\ccsdesc[500]{Computing methodologies~Transfer learning}
\ccsdesc[500]{Computing methodologies~Neural networks}
\ccsdesc[500]{Security and privacy~Intrusion detection systems}

\keywords{Domain Adaptation, Internet of Things, Intrusion Detection, Open-Set Domain Adaptation, Dandelion Network}

\received{28 March 2023}
\received[revised]{21 September 2023}
\received[accepted]{4 January 2024}

\maketitle

\section{Introduction}\label{sec:introduction}

Internet of Things (IoT) devices become prevalent in many real-world applications \cite{10049541,9833301,8764459,10.1145/3394504}. However, they tend to be computational and energy-constrained, which hinder the deployment of effective intrusion detection mechanisms. Together with the lack of maintenance, these limitations compromise the security of IoT devices, making them vulnerable to attacks \cite{lu2018internet,9933783}. To protect the safety of IoT devices, an effective intrusion detection mechanism becomes indispensable \cite{10026337}. 

The intrusion detection for IoT has drawn wide attention from the academic community. For instance, signature-based intrusion detectors were proposed \cite{mitchell2013behavior,8999496,mitchell2013adaptive}, which detected malicious behaviours by pattern matching with sophisticated rule repositories. With the rapid growth of machine learning techniques, some machine learning and deep learning-based intrusion detectors were also proposed \cite{muhammad2020stacked,murali2019lightweight,yao2019capsule} and achieved satisfactory performance. However, these traditional intrusion detection methods \rw{either require} a sophisticated, thorough and up-to-date rule repository, or a fully annotated training dataset. These prerequisites either require comprehensive expertise knowledge to build and update, or require a tremendous \rw{amount of efforts} to annotate. Besides, due to the limited storage and communication capability of the IoT device and the concerns of user privacy, it further hinders the availability of an IoT intrusion rule repository or training dataset. Under such data-scarcity \cite{10026337}, these traditional intrusion detectors suffer from compromised performance. 

To work around the data-scarcity, domain adaptation-based (DA) intrusion detection methods \cite{10.1145/3394171.3413995} can be leveraged by transferring the intrusion knowledge from a knowledge-rich source network intrusion (NI) domain to assist the intrusion detection for the target IoT intrusion (II) domain. Popular solutions \cite{10026337,9933783} performed intrusion knowledge transfer and meanwhile masked the heterogeneities between different domains and achieved satisfying outcomes. 

Despite the effectiveness of these DA-based methods, they operate under the assumption that both source and target domains share exactly the same type of intrusions. However, this assumption is sometimes unrealistic in the real-world as the IoT intrusion domain can constantly confront newly-emerged intrusion strategies \cite{mehedi2022dependable}. Therefore, this assumption hinders the applicability of traditional DA-based intrusion detectors. As a more general solution, Open-Set Domain Adaptation (OSDA) \cite{panareda2017open,fang2020open} relaxes this assumption and allows the \rw{target domain} to contain newly-emerged intrusions unobserved in the source NI domain. Some OSDA methods were proposed \cite{li2022interpretable,jing2021towards,luo2020progressive}, which tackled this challenging setting via hyperspherical feature space learning, semantic recovery learning and progressive graph learning, etc. However, these research efforts all suffered from some drawbacks, such as failing to utilise the graph embedding alignment in the learned hyperspherical feature space, lacking the exploitation of the correction effect of the semantics, etc., which therefore provided room for improvement of a more effective OSDA-based intrusion detector. 

In this paper, inspired by the structure of the dandelion, we propose the Open-Set Dandelion Network (OSDN) based on unsupervised heterogeneous DA in an open-set manner. The OSDN model tackles the IoT data scarcity by transferring intrusion knowledge from a knowledge-rich source NI domain to assist the knowledge-scarce IoT target domain. It relaxes the closed-set assumption and can effectively detect both known and unknown intrusions faced by IoT devices, making it applicable in real-world applications. To achieve this, the OSDN model forms the source domain into a dandelion-like feature space with the goal of \rw{grouping each intrusion category compactly and meanwhile separating different intrusion categories, i.e., } achieving inter-category separability and intra-category compactness, the foundation for an accurate intrusion detector to work on. The dandelion-based target membership mechanism then constructs the target dandelion. Then, the dandelion angular separation mechanism is leveraged to enhance inter-category separability, together with the dandelion embedding alignment mechanism, which transfers intrusion knowledge via a graph embedding perspective. The discriminating sampled dandelion mechanism is also used to promote intra-category compactness. Besides, trained using both known and generated unknown intrusion knowledge, the intrusion classifier produces probabilistic semantics, which forms a semantic dandelion and in turn emphasises easily-confused categories and provide correction for better inter-category separability. Holistically, these mechanisms form the OSDN model that can effectively transfer intrusion knowledge for more accurate IoT intrusion detection. 

In summary, the contributions of this paper are three-fold as follows: 
\begin{itemize}
  \item We realise the benefits of the Open-Set DA technique to perform intrusion knowledge transfer and facilitate more accurate intrusion detection for the data-scarce IoT scenarios. The OSDA-based intrusion detector also relaxes the closed-set assumption, making it a more robust intrusion detector in the real-world. 
  \item We formulate the intrusion feature space into a dandelion-like feature space. The proposed OSDN model leverages mechanisms such as the dandelion angular separation mechanism (DASM), the dandelion embedding alignment mechanism (DEAM), the discriminating sampled dandelion mechanism (DSDM) and the semantic dandelion correction mechanism (SDCM) to promote inter-category separability and intra-category compactness in the dandelion feature space, which is the foundation for an accurate intrusion detector to work on. 
  \item We conduct comprehensive experiments on five widely recognised intrusion detection datasets and verify the effectiveness of the OSDN model against three state-of-the-art baselines. A $16.9\%$ performance boost is achieved. Besides, the contribution of each OSDN constituting component, the stability and the efficiency of the OSDN model is also verified. 
\end{itemize}

The rest of the paper is organised as follows: Section \ref{sec:related_work} categorises related works and summarises the research opportunities. Section \ref{sec:model_preliminary_and_architecture} presents model preliminaries and the OSDN model architecture, followed by Section \ref{sec:the_osdn_algorithm}, in which the detailed mechanisms constituting the OSDN model are presented. Section \ref{sec:experiment} presents the experimental setup and detailed experimental analyses. The last section concludes the paper. We provide an acronym table and a notation table for better readability in Appendix \ref{sec:appendix}. 

\section{Related Work}\label{sec:related_work}

\begin{figure*}[!ht]
  \begin{center}
    \includegraphics[width=0.7\textwidth,keepaspectratio]{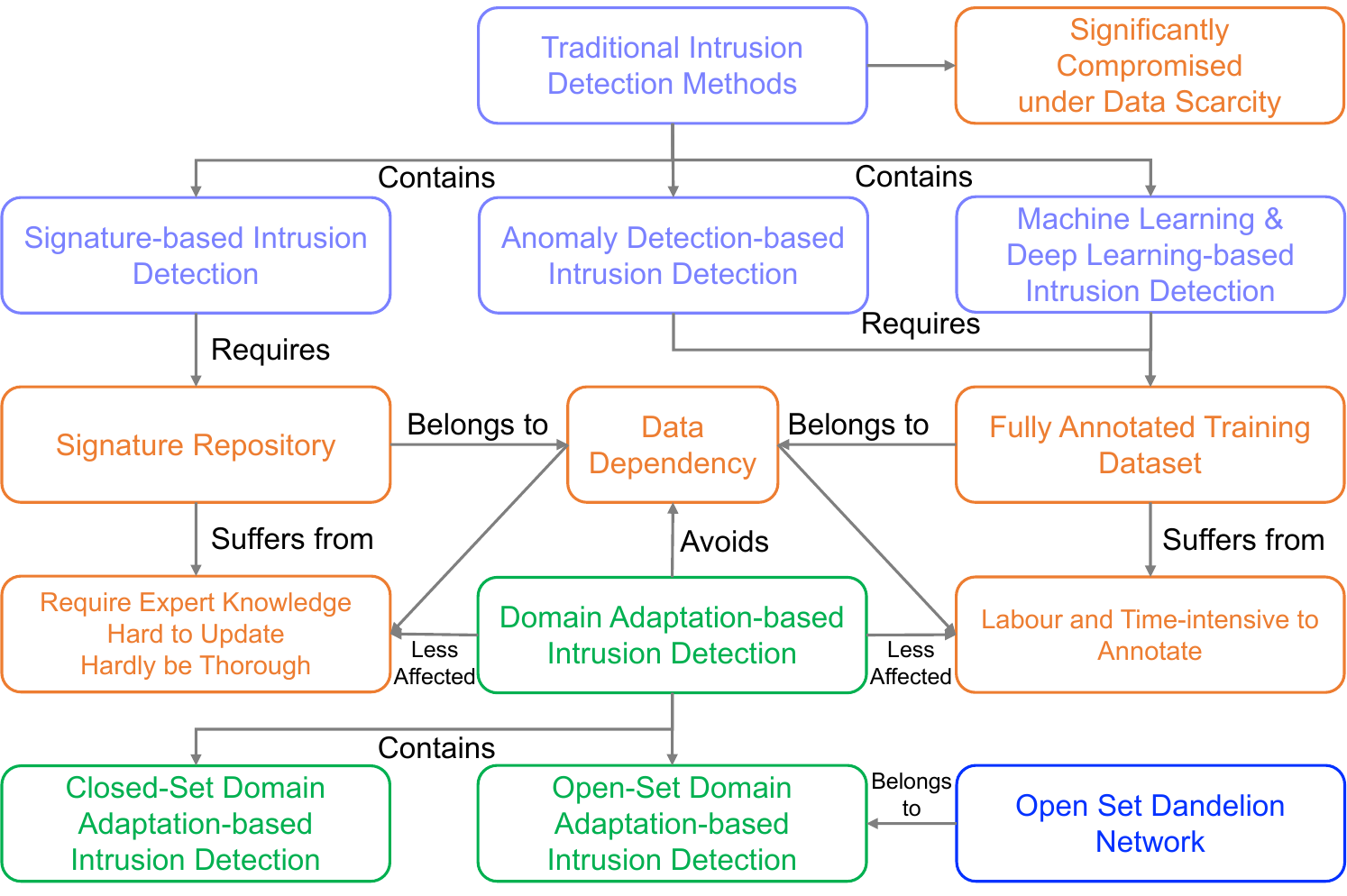}\\
    \caption{\rw{Summarisation of IoT intrusion detection methods, the data dependency and drawback of traditional intrusion detection methods, and the merits of domain adaptation-based intrusion detection methods. The OSDN method belongs to the open-set domain adaptation-based intrusion detector. }}
    \vspace{-4mm}
    \label{fig:figure_ids_methods}
  \end{center}
\end{figure*}

In this section, we introduce the related works in a categorised manner and outline our research opportunities. \rw{In Fig. \ref{fig:figure_ids_methods}, we summarise the traditional IoT intrusion detection methods, their data dependency and their drawbacks, which reflect the merits of the domain adaptation-based intrusion detection methods for the data-scarce IoT scenarios. The OSDN method belongs to the open-set domain adaptation-based intrusion detector. }

\subsection{Traditional Intrusion Detection}\label{sec:traditional_intrusion_detection}

Intrusion detection has drawn wide attention from the research community. Traditional intrusion detection methods, including signature-based intrusion detectors \cite{mitchell2013behavior,mitchell2013adaptive,dietz2018iot}, \rw{which require a sophisticated rule repository for decision making. }It can only detect malicious intrusions if their patterns match certain rules in the repository. \rw{Anomaly-based intrusion detectors \cite{bovenzi2023network,satam2020wids,bovenzi2020hierarchical,tavallaee2010toward} are also popular. }These methods need to go through a comprehensive training process \rw{based on a well annotated training dataset} to learn the patterns of normal traffic behaviours and then flag any traffic that deviates from the normal patterns. With the rapid advance of machine learning and deep learning techniques, ML and DL-based intrusion detectors are also widely used. Possible methods include multi-kernel SVM \cite{murali2019lightweight}, isolation forest \cite{eskandari2020passban} and \rw{deep learning models such as autoencoders \cite{muhammad2020stacked,mirsky2018kitsune}} and capsule network \cite{yao2019capsule}, etc. 

However, all these traditional intrusion detection methods may be hindered by the IoT data-scarcity \rw{due to their strong data dependency on a well-built intrusion rule repository or a finely-annotated training dataset}. \rw{Building an intrusion rule repository requires sophisticated expertise knowledge, and can hardly be thorough and up-to-date. Besides, finely annotating a training dataset is both labour and time-intensive. Without enough annotated datasets, the learning process of anomaly-based, ML-based and DL-based methods is significantly hindered, resulting in compromised efficacy. }Therefore, it naturally leads to the domain adaptation-based solutions, which can work under data-scarce IoT scenarios by performing intrusion knowledge transfer\rw{, a merit that outperforms traditional intrusion detection methods. }

\subsection{Domain Adaptation for Intrusion Detection}\label{sec:domain_adaptation_for_intrusion_detection}

Domain adaptation can transfer intrusion knowledge from a knowledge-rich source domain to facilitate more accurate intrusion detection for the target domain. Hence, \rw{it possesses the merit to comfortably work under the data-scarce IoT scenario. }Wu et al., \cite{9933783} proposed a Joint Semantic Transfer Network, aiming to address the IoT intrusion detection problem under the semi-supervised heterogeneous DA setting. Later, the Geometric Graph Alignment method was also proposed by Wu et al., \cite{10026337} to tackle the intrusion detection for completely unsupervised target IoT domains. There are other DA methods such as \cite{xie2022collaborative,liang2021pareto,10082914}, which performed intrusion knowledge transfer via Wasserstein distance minimisation, adversarial learning, Pareto optimal solution searching, adaptive recommendation matching, etc. 

However, the traditional domain adaptation methods work under the closed-set assumption that the intrusion categories in both source and target domains are exactly the same. Hence, these methods cannot tackle the case in which new IoT intrusions emerge as time goes by, limiting their applicability in the real-world. 

\subsection{Open-Set Domain Adaptation for Intrusion Detection}\label{open_set_domain_adaptation_for_intrusion_detection}

Open-Set DA methods relax the closed-set assumption of traditional DA methods and allow the target IoT domain to possess new intrusions unobserved in the source domain. Jing et al., \cite{jing2021towards} presented an open-set DA method with semantic recovery to better exploit the semantic information of the unknown target intrusions. However, it put no effort to explore the possibility brought by the hyperspherical structure formulation with excellent inter-category distinguishability. Li et al., \cite{li2022interpretable} explored the open-set DA problem via the angular margin separation network. Despite its effectiveness, it lacked finer alignment achievable by graph embedding and ignored the correction effect of the semantics. Besides, Luo et al., \cite{luo2020progressive} investigated the graph embedding-based open-set DA solution. However, the proposed Progressive Graph Learning (PGL) also failed to investigate the usefulness of angular-based hyperspherical space with excellent separability and compactness. 

\subsection{Research Opportunity}\label{sec:research_opportunity}

The OSDN model transfers intrusion knowledge via the dandelion-based feature space that emphasises both inter-category separability and intra-category compactness, which is lacked by previous open-set DA methods as in \cite{jing2021towards,luo2020progressive}. Besides, the graph embedding alignment can achieve both finer feature space alignment and tighter intra-category structure via adversarial learning. Such mechanisms were not attempted in \cite{jing2021towards,li2022interpretable}. Moreover, the OSDN model leverages the semantic dandelion correction mechanism, the utilisation of the semantic dandelion fills the void in \cite{li2022interpretable}. The semantic correction is also lacked in these aforementioned methods. By combining these methods to form a holistic framework, the OSDN model can perform finer intrusion knowledge transfer and benefit IoT intrusion detection. 

\section{Model Preliminary and Architecture}\label{sec:model_preliminary_and_architecture}

In this section, we introduce the preliminaries and the architecture of the proposed OSDN model. 

\subsection{Model Preliminary}\label{sec:model_preliminary}

The OSDN model works under the unsupervised open-set DA setting with heterogeneities exist between domains. Following common notations in \cite{9933783}, we denote the source NI domain $\mathcal{D}_{S}$ as follows: 
\begin{equation}
  \begin{split}
    & \mathcal{D}_{S} = \{\mathcal{X}_{S}, \mathcal{Y}_{S}\} = \{(x_{S_i}, y_{S_i})\}, i \in [1, n_{S}], \\
    & x_{S_i} \in \mathbb{R}^{d_{S}}, y_{S_i} \in [1, K]\,,
  \end{split}
\end{equation}
where $\mathcal{X}_{S}$ contains $n_{S}$ source NI domain traffic features, each feature vector is represented in $d_{S}$ dimensions. $\mathcal{Y}_{S}$ is the corresponding intrusion category label within a total number of $K$ categories, one normal category and others are intrusion categories. Similarly, the target II domain $\mathcal{D}_{T}$ is defined as follows: 
\begin{equation}
  \begin{split}
    & \mathcal{D}_{T} = \{\mathcal{X}_{T}, \mathcal{Y}_{T}\} = \{(x_{T_i}, y_{T_i})\}, i \in [1, n_{T}], \\
    & x_{T_i} \in \mathbb{R}^{d_{T}}, y_{T_i} \in [1, K'], K' > K\,.
  \end{split}
\end{equation}
Under the open-set DA setting, the intrusion categories of the source NI domain is a subset of the intrusion categories of the target II domains, i.e., $\mathcal{Y}_{S} \subset \mathcal{Y}_{T}$, $K' > K$. Both domains share $K$ common intrusion categories. Furthermore, the target II domain contains $K' - K$ new intrusion categories unobserved in the source domain. Under the unsupervised setting, the ground truth labels of the target II domain remain agnostic during the training process. As a heterogeneous DA problem, heterogeneities present between domains, e.g., $d_{S} \neq d_{T}$. 

\subsection{The OSDN Architecture}\label{sec:the_osdn_architecture}

\begin{figure*}[!ht]
  \begin{center}
    \includegraphics[width=\textwidth,keepaspectratio]{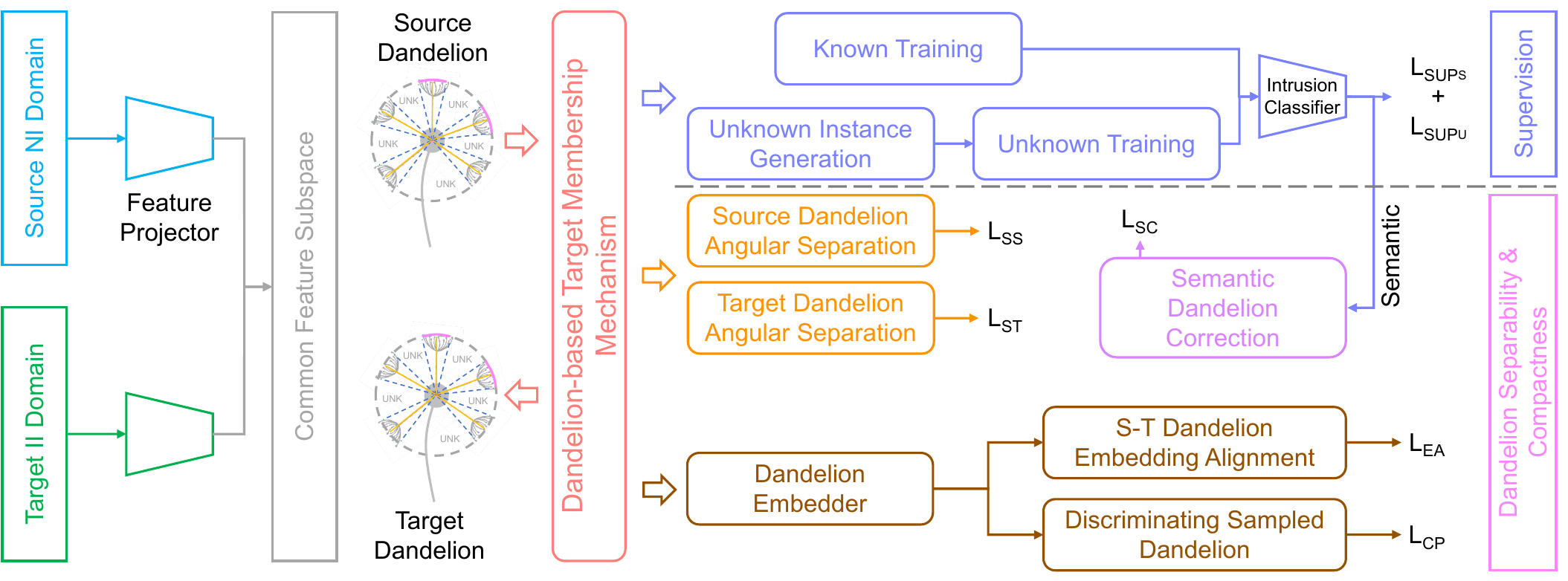}\\
    \caption{The architecture of the OSDN model and the interrelationships between the OSDN's constituting components. }
    \vspace{-4mm}
    \label{fig:figure_architecture}
  \end{center}
\end{figure*}

\begin{figure*}[!ht]
  \begin{center}
    \includegraphics[width=0.66\textwidth,keepaspectratio]{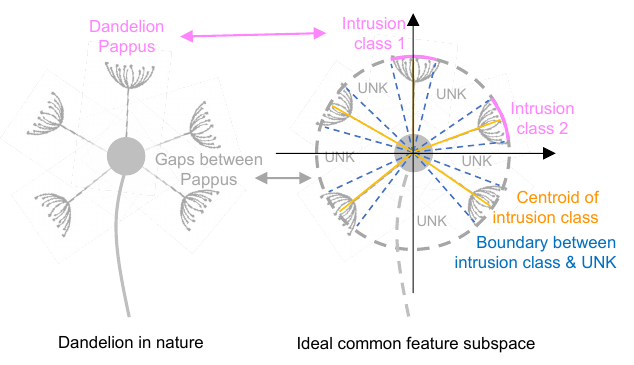}\\
    \caption{The analogy between the structure of a dandelion and the dandelion-like common feature subspace. Each pappus corresponds to a shared intrusion category, it needs to be compact and well-separated from other intrusion categories (pappuses)\rw{, simultaneously achieving both intra-category compactness and inter-category separability}. Target unknown intrusion categories reside in the gaps between pappuses to achieve distinguishability. The analogy between dandelion and the ideal common feature subspace leads to the naming of the Open-Set Dandelion Network (OSDN). }
    \vspace{-4mm}
    \label{fig:figure_dandelion}
  \end{center}
\end{figure*}

The architecture of the OSDN model has been presented in Fig. \ref{fig:figure_architecture}. To perform intrusion knowledge transfer, features in each domain will be normalised to form a unit hyperspherical space and then be projected into a $d_{C}$-dimensional common feature subspace (the grey box) by its corresponding feature projector (the trapezoids). The feature projector $E$ is defined as follows: 
\begin{equation}
  \begin{split}
    & f(x_i) = \begin{cases}
      E_{S}(x_i) & \text{if $x_i \in \mathcal{X}_S$} \\
      E_{T}(x_i) & \text{if $x_i \in \mathcal{X}_T$}
    \end{cases} \\
    & f(x_i) \in \mathbb{R}^{d_{C}}\,.
  \end{split}
\end{equation}
As illustrated in Fig. \ref{fig:figure_dandelion}, the common feature subspace aims to group each shared intrusion category in a compact manner (each pappus of the dandelion\rw{, i.e., intra-category compactness}), and meanwhile achieves excellent separability between intrusion categories\rw{, i.e., inter-category separability}. For these unknown new intrusions in the target II domain, since their number is agnostic, therefore, instead of deliberately grouping them in a brute-force manner, the dandelion-analogous common feature subspace allows them to spread in any gap between pappuses to promote distinguishability between shared and unknown intrusion categories. As visualised in Fig. \ref{fig:figure_dandelion}, by making the common feature subspace analogous to the structure of the dandelion, i.e., achieving excellent intra-category compactness and inter-category separability, the shared classifier $C$ can then make accurate intrusion detection decisions. 

The source dandelion can be formed directly since the source domain is completely supervised. Then, the dandelion-based target membership mechanism (the red box in Fig. \ref{fig:figure_architecture}) is used to form the target dandelion based on the spatial relationship between target instances and the source dandelion. Once both source and target dandelions are formed, the dandelion angular separation mechanism (the orange boxes in Fig. \ref{fig:figure_architecture}) is utilised to enhance the inter-category separability in each dandelion. Besides, a dandelion embedder is leveraged to generate graph embeddings for dandelions and it is used in two ways (the brown boxes in Fig. \ref{fig:figure_architecture}): the graph embeddings of source and target dandelions are aligned to promote better alignment between domains; moreover, sampled child dandelions are produced and their graph embeddings need to confuse a discriminator to achieve finer intra-category compactness. To better train the shared intrusion classifier $C$, the source domain data provides supervision information. To equip the intrusion classifier with knowledge of target unknown intrusions, unknown instances residing in the pappus gaps in the source dandelion are generated for unknown intrusion training. Lastly, the probabilistic semantic yielded by the shared intrusion classifier also works as a correction to deliberately emphasise easily-confused categories and remind the dandelion angular separation mechanism to separate them, forming a correction loop (the purple box in Fig. \ref{fig:figure_architecture}). 

Finally, by forming these mechanisms into a holistic model, fine-grained intrusion knowledge transfer can be achieved and the shared and unknown intrusion categories will be well-separated so that the shared classifier $C$ can enjoy excellent intrusion detection efficacy for the target II domain.

\section{The OSDN Algorithm}\label{sec:the_osdn_algorithm}

In this section, we present the detailed mechanism of each OSDN constituting component and the overall optimisation objective of the model. 

\subsection{Dandelion-based Target Membership Mechanism (DTMM)}\label{sec:dandelion_based_target_membership_mechanism}

The source dandelion can be easily formed based on its supervision information. Then, the source dandelion will guide the membership decision for unsupervised target instances to form the target dandelion. For each source intrusion category $i$, the maximum intra-category deviation $d_{max}^{i}$ will be calculated as follows: 
\begin{equation}\label{equ:maximum_intra_category_deviation}
  \begin{split}
    d_{max}^{(i)} &= \text{max}(1 - \text{COS}(x_{S_j}^{(i)}, \mu_{S}^{(i)})), j \in [1, n_{S}^{(i)}], \\
    \mu_{S}^{(i)} &= \frac{1}{n_{S}^{(i)}} \sum_{j=1}^{n_{S}^{(i)}} x_{S_j}^{(i)}\,,
  \end{split}
\end{equation}
where $COS()$ stands for Cosine Similarity, $n_{S}^{(i)}$ denotes the number of instances in the $i$\textsuperscript{th} intrusion category in the source domain, $\mu_{S}^{(i)}$ denotes the mean of the source intrusion category $i$ and $x_{S_j}^{(i)}$ means the $j$\textsuperscript{th} instance of the source $i$\textsuperscript{th} intrusion category. Then, each target instance will be assigned to its nearest source category $i$ if it resides within the maximum deviation range of source category $i$, i.e., 
\begin{equation}
  \begin{split}
    y_{T_j}^{D} &= \begin{cases}
      \underset{i}{\mathrm{argmin}} (1 - COS(x_{T_j}, \mu_{S}^{(i)})) & \text{if $1 - COS(x_{T_j}, \mu_{S}^{(i)}) \leq d_{max}^{(i)}$} \\
      K+1 & \text{otherwise}\,,
    \end{cases}
  \end{split}
\end{equation}
where $y_{T_j}^{D}$ represents the dandelion-based membership for the $j$\textsuperscript{th} target instance $x_{T_j}$. Otherwise, that target instance will be assigned to the unknown category $K+1$ to avoid deteriorating the compactness of its closest intrusion category. Unlike methods such as \cite{jing2021towards,li2022interpretable} that perform K-means clustering of unknown intrusions, the OSDN assigns all unknown intrusions into a single category $K+1$ and hence does not rely on the availability of the prior knowledge on the number of unknown intrusion categories and is more practical in the real-world. Besides, the OSDN model does not deliberately enforce all unknown target instances to reside at a single place, it allows unknown intrusions to reside at any pappus gap in the target dandelion. Deliberately aligning unknown target instances coming from different intrusion categories may cause negative transfer. 

\subsection{Dandelion Angular Separation Mechanism (DASM)}\label{sec:dandelion_angular_separation_mechanism}

To increase the separability between known intrusion categories and meanwhile enhance the discriminability between known and unknown intrusion categories, i.e., enlarge the gap between pappuses, the OSDN model will achieve these goals from an angular perspective. First, the centroid of each intrusion category will be calculated. Then, the source category pair-wise Cosine similarity matrix $CS_{S}$ will be calculated as follows: 
\begin{equation}
  \begin{split}
    CS_{S} &= \begin{bmatrix}
      CS_{S}^{11} & \boldsymbol{CS_{S}^{12}} & \boldsymbol{\cdots} & \boldsymbol{CS_{S}^{1K}}\\
      CS_{S}^{21} & CS_{S}^{22} & \boldsymbol{\cdots} & \boldsymbol{CS_{S}^{2K}}\\
      \vdots & \vdots & \ddots & \boldsymbol{\vdots}\\
      CS_{S}^{K1} & CS_{S}^{K2} & \cdots & CS_{S}^{KK}
    \end{bmatrix}\,,\\
    CS_{S}^{ij} &= COS(\mu_{S}^{(i)}, \mu_{S}^{(j)})\,,
  \end{split}
\end{equation}
where $CS_{S}^{ij}$ represent the Cosine similarity between the $i$\textsuperscript{th} and $j$\textsuperscript{th} intrusion category of the source NI domain. By minimising the sum of the upper triangle of the matrix $CS_{S}$, it enlarges the inter-category angular divergence. The source dandelion separation loss $\mathcal{L}_{SS}$ is defined as follows: 
\begin{equation}\label{equ:source_dandelion_separation_loss}
  \mathcal{L}_{SS} = \frac{2}{K(K-1)} \sum_{i=1}^{K-1} \sum_{j=i+1}^{K} CS_{S}^{ij}\,.
\end{equation}
The target dandelion Cosine similarity matrix $CS_{T}$ and the corresponding target dandelion separation loss $\mathcal{L}_{ST}$ are defined similarly. By minimising both $\mathcal{L}_{SS}$ and $\mathcal{L}_{ST}$, it promotes better dandelion inter-category separability from an angular perspective. 

\subsection{Dandelion Embedding Alignment Mechanism (DEAM)}\label{sec:dandelion_embedding_alignment}

To further promote a finer alignment between the source and target dandelions, a dandelion graph embedder is used to produce the graph embeddings for both dandelions. To achieve this, each dandelion is formulated as a graph, defined as follows: 
\begin{equation}
  \begin{split}
    G_{S} &= <V_{S}, E_{S}>\\
    V_{S} &= \{V_{S}^{(i)}\}, i \in [1, K], V_{S}^{(i)} = \mu_{S}^{(i)}\\
    E_{S} &= \{E_{S}^{i,j}\}, E_{S}^{i,j} = ||\mu_{S}^{(i)} - \mu_{S}^{(j)}||_2^2, i \in \{\vmathbb{0}\} \cup [1, K], j \in \{\vmathbb{0}\} \cup [1, K], i \neq j\,,
  \end{split}
\end{equation}
where $G_{S}$ denotes the source dandelion graph, $V_{S}$ and $E_{S}$ stand for vertices and edges in $G_{S}$, respectively and $G_{T}$ is defined similarly. Each vertex $V_{S}^{(i)}$ is the centroid of the corresponding intrusion category. The graph is fully connected and each vertex is also connected with the origin, denoted as $\vmathbb{0}$. 

In the OSDN model, we apply the Feather network \cite{10.1145/3340531.3411866} as the graph embedder. As a graph embedding algorithm, it enjoys several merits: first, the Feather network can work in an unsupervised manner, which works comfortably under the data-scarce IoT scenario; second, the Feather network enjoys a linear time complexity as proved in \cite{10.1145/3340531.3411866}, the low complexity can enhance the efficiency of the intrusion detection model in real-world applications; finally, the Feather network is comprehensively verified \cite{10.1145/3340531.3411866} to have superior graph embedding performance. 

Using the graph embedder, each dandelion graph will be mapped into a $d_{G}$-dimensional graph embedding space, in which the more geometrically similar between dandelion graphs, the more similar the graph embeddings will be. Then, the dandelion embedding alignment loss $\mathcal{L}_{EA}$ is defined as follows: 
\begin{equation}
  \mathcal{L}_{EA} = ||\phi_{S} - \phi_{T}||_{2}^{2}, \phi_{S}, \phi_{T} \in \mathbb{R}^{d_{G}}\,,
\end{equation}
where $\phi_{S}$ denotes the graph embedding of the source domain dandelion. By minimising the dandelion embedding alignment loss, both dandelions will be further aligned and hence will promote better intrusion knowledge transfer, as verified by experimental evidences in Section \ref{sec:ablation_study}. 

\subsection{Discriminating Sampled Dandelion Mechanism (DSDM)}\label{sec:discriminating_sampled_dandelion_mechanism}

\begin{figure*}[!ht]
  \begin{center}
    \includegraphics[width=0.75\textwidth,keepaspectratio]{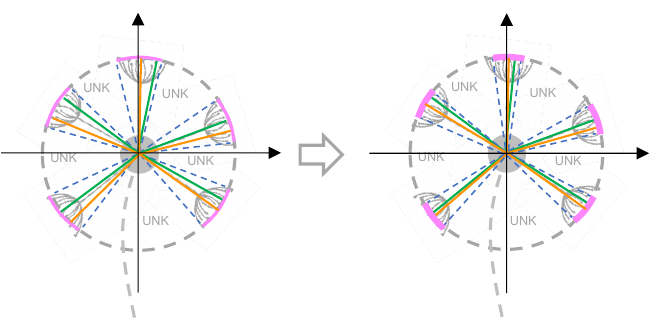}\\
    \caption{Illustrating example of the OSDN discriminating sampled dandelion mechanism to enhance intra-category compactness. }
    \vspace{-4mm}
    \label{fig:figure_discriminating_compactness}
  \end{center}
\end{figure*}

To further boost the intra-category compactness and hence promote better known-intrusion separability and unknown-intrusion discriminability, a discriminating sampled dandelion mechanism is proposed. As illustrated in Fig. \ref{fig:figure_discriminating_compactness}, one instance per intrusion category is randomly sampled to form a new child dandelion, such as the orange and the green dandelion in Fig. \ref{fig:figure_discriminating_compactness}. The more compact each intrusion category is, the more similar the embeddings of child dandelions will be. Hence, the OSDN achieves this goal via a discriminating perspective. First, both source and target domain intrusion features will be fused in the common feature subspace to form a fused dandelion, then, $N$ child dandelions will be sampled, where the $i$\textsuperscript{th} pappus in each child dandelion is a randomly selected instance from the $i$\textsuperscript{th} category from the fused dandelion. Next, a discriminator is confused using the discriminating sampled dandelion loss $\mathcal{L}_{CP}$, defined as follows: 
\begin{equation}
  \mathcal{L}_{CP} = \frac{1}{2} (log(D(\phi_{S})) + log(D(\phi_{T}))) + \frac{1}{N} \sum_{j=1}^{N} (1 - log(D(\phi_{\mathcal{DD}_{*}}^{j})))\,,
\end{equation}
in which $\mathcal{DD}_{S}$, $\mathcal{DD}_{T}$ and $\mathcal{DD}_{*}^{j}$ denote the source, target and $j$\textsuperscript{th} sampled dandelion, $\phi$ denotes the dandelion graph embedding and $D()$ denotes the discriminator. By assigning $\mathcal{DD}_{S}$ and $\mathcal{DD}_{T}$ with label $1$ and assign sampled child dandelions with label $0$, letting the network to minimise the $\mathcal{L}_{CP}$ will confuse the discriminator to be incapable to distinguish whether the given dandelion embedding is generated from a randomly sampled dandelion or not. Meanwhile, the discriminator will try to stay unconfused. Once the minimax game between the network and the discriminator reaches an equilibrium, the graph embeddings of source, target and sampled child dandelions will become indistinguishable, which in turn enhances the intra-category compactness, as illustrated in Fig. \ref{fig:figure_discriminating_compactness}. 

\subsection{Semantic Dandelion Correction Mechanism (SDCM)}\label{sec:semantic_dandelion_correction}

\begin{figure*}[!ht]
  \begin{center}
    \includegraphics[width=0.66\textwidth,keepaspectratio]{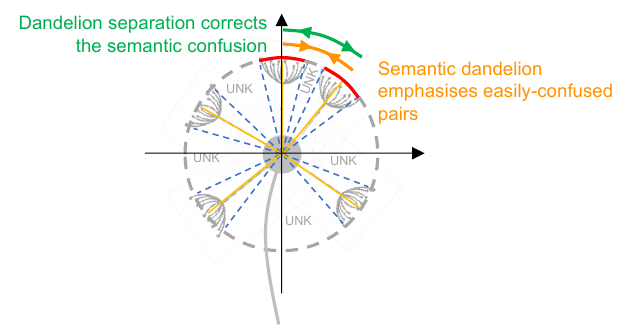}\\
    \caption{The OSDN semantic dandelion correction mechanism. It will point out easily confused intrusion category pairs from the probabilistic semantic perspective (the orange part), which will act as a correction to the dandelion angular separation mechanism (the green part). }
    \vspace{-4mm}
    \label{fig:figure_semantic_correction}
  \end{center}
\end{figure*}

The source NI domain is completely supervised, however, it lacks the knowledge of unknown intrusions in the target II domain. Therefore, directly using the source NI domain supervision to train the shared intrusion classifier $C$ will significantly hinder its ability to detect unknown intrusions. To tackle this issue, the OSDN model generates $n_{R}$ instances residing in the gaps between source dandelion pappuses, and treat these generated instances as unknown intrusions to equip the intrusion classifier $C$ with the ability to detect both known and unknown intrusions under the open-set DA setting. The overall supervision loss of known and unknown training $\mathcal{L}_{SUP}$ is defined as follows: 
\begin{equation}
  \begin{split}
    \mathcal{L}_{SUP} &= \mathcal{L}_{SUP_{S}} + \mathcal{L}_{SUP_{U}}\\
     &= \frac{1}{n_{S}} \sum_{j=1}^{n_{S}} \mathcal{L}_{CE}(C(f(x_{j})), y_{j}) + \frac{1}{n_{R}} \sum_{j=1}^{n_{R}} \mathcal{L}_{CE}(C(f(x_{j})), y_{j})\\
    y_{j} &= \begin{cases}
      y_{S_j} & \text{if $x_{j} \in \mathcal{X}_{S}$}\\
      K+1 & \text{if $x_{j} \in \mathcal{X}_{R}$}
    \end{cases}\,,
  \end{split}
\end{equation}
where $\mathcal{L}_{CE}$ denotes the cross entropy loss and $\mathcal{X}_{R}$ represents generated unknown instances for unknown training. 

Once the intrusion classifier $C$ is well-trained, it can then yield probabilistic semantics for each intrusion data instance $j$, i.e., the inter-category probabilistic correlations, denoted as $p_{j}$. Therefore, the semantic information can also form new semantic dandelions $\mathcal{DD}_{S*}$ in the semantic space, defined as follows
\begin{equation}
  \mathcal{DD}_{\mathcal{S}S}^{(i)} = \frac{1}{n_{S}^{(i)}} \sum_{j=1}^{n_{S}^{(i)}} p_{S_j}^{(i)}, \mathcal{DD}_{\mathcal{S}T}^{(i)} = \frac{1}{|y_{T}^{D}=i|} \sum_{j=1}^{|y_{T}^{D}=i|} p_{T_j}^{(i)}\,,
\end{equation}
where $\mathcal{DD}_{\mathcal{S}S}^i$ denotes the $i$\textsuperscript{th} pappus of the source semantic dandelion $\mathcal{DD}_{\mathcal{S}S}$, $n_{S}^{(i)}$ represents the number of source $i$\textsuperscript{th} category instances, $y_{T}^{D}$ denotes the membership assigned to target instances by the source dandelion in Section \ref{sec:dandelion_based_target_membership_mechanism}. Then, the Cosine similarity matrix $CS_{SM}$ between both semantic dandelions are calculated as follows: 
\begin{equation}
  \begin{split}
    CS_{SM} &= \begin{bmatrix}
      \boldsymbol{CS_{SM}^{11}} & \boldsymbol{CS_{SM}^{12}} & \boldsymbol{\cdots} & \boldsymbol{CS_{SM}^{1K}}\\
      CS_{SM}^{21} & \boldsymbol{CS_{SM}^{22}} & \boldsymbol{\cdots} & \boldsymbol{CS_{SM}^{2K}}\\
      \vdots & \vdots & \boldsymbol{\ddots} & \boldsymbol{\vdots}\\
      CS_{SM}^{K1} & CS_{SM}^{K2} & \cdots & \boldsymbol{CS_{SM}^{KK}}
    \end{bmatrix}\,,\\
    CS_{SM}^{ij} &= COS(\mathcal{DD}_{\mathcal{S}S}^{(i)}, \mathcal{DD}_{\mathcal{S}T}^{(j)})\,.
  \end{split}
\end{equation}
Ideally, the $i$\textsuperscript{th} intrusion category from both source NI and target II domain should share similar inter-category probabilistic semantics, while different intrusion categories from both domains should have their inter-category probabilistic semantics diverge from each other. To achieve this, the OSDN model minimises the semantic dandelion correction loss $\mathcal{L}_{SC}$ as follows: 
\begin{equation}
  \mathcal{L}_{SC} = \frac{2}{K(K+1)} \sum_{i=1}^{K} \sum_{j=i}^{K} CS_{SM}^{ij}\,.
\end{equation}
By minimising the $CS_{SM}^{ij}, i \neq j$, inter-category probabilistic semantics will be diverged from each other, leading to better inter-category discriminability. It is worth noting that the $\mathcal{L}_{SC}$ also minimises the $CS_{SM}^{ii}$, i.e., maximising the divergence between cross-domain same-category probabilistic semantics. The rationale is as follows: if minimising the $CS_{SM}^{ii}$ can easily compromise the semantic of the $i$\textsuperscript{th} intrusion category, then it indicates the $i$\textsuperscript{th} intrusion category can be easily confused with other categories from the probabilistic semantic perspective, as indicated in Fig. \ref{fig:figure_semantic_correction}. Therefore, deliberately minimising $CS_{SM}^{ii}$ can exploit and emphasise easily-confused intrusion category pairs, i.e., pointing out a possible point to correct for the dandelion angular separation mechanism. Consequently, by utilising this correction mechanism, it can further boost the dandelion separation efficacy, as supported by experimental evidences in Section \ref{sec:ablation_study} and in turn enhances the intrusion detection accuracy. 

\subsection{Overall Optimisation Objective}\label{sec:overall_optimisation_objective}

Overall, the optimisation objective of the OSDN model is defined as follows: 
\begin{equation}
  \begin{split}
    & \min_{E_{S}, E_{T}, C} (\alpha_{S} \mathcal{L}_{SUP_{S}} + \alpha_{U} \mathcal{L}_{SUP_{U}} + \beta_{S} \mathcal{L}_{SS} + \beta_{T} \mathcal{L}_{ST} + \delta \mathcal{L}_{EA} + \theta \mathcal{L}_{SC} + \gamma \mathcal{L}_{CP})\\
    & \max_{D} (\gamma \mathcal{L}_{CP})\,,
  \end{split}
\end{equation}
where $\alpha_{S}$, $\alpha_{U}$, $\beta_{S}$, $\beta_{T}$, $\delta$, $\theta$ and $\gamma$ are hyperparameters controlling the influence of the corresponding loss components. We utilise the gradient reversal layer \cite{ganin2016domain} for the discriminator, which acts as an identity function during forward propagation and reverses the gradient during backpropagation to achieve an end-to-end optimisation process for the OSDN model. Once the above minimax game reaches an equilibrium, the intrusion knowledge is transferred in a fine-grained manner, and the intrusion detection efficacy can therefore benefit.

\section{Experiment}\label{sec:experiment}

To verify the effectiveness of the OSDN model, we perform experiments on five comprehensive and representative intrusion detection datasets with three state-of-the-art baseline counterparts. We also verify the performance stability of the OSDN model under varied openness settings and manipulated hyperparameter settings and demonstrate the contribution and necessity of each OSDN constituting component. Finally, we verify the computational efficiency of the OSDN model. 

\subsection{Experimental Datasets}\label{sec:experimental_datasets}

We use five comprehensive intrusion detection datasets. Network intrusion detection datasets includes NSL-KDD, UNSW-NB15 and CICIDS2017. IoT intrusion detection datasets includes UNSW-BOTIOT and UNSW-TONIOT. 

\textbf{Network Intrusion Dataset: NSL-KDD} This dataset \cite{tavallaee2009detailed} contains benign network traffic and four types of real-world intrusions, such as probing attacks, Denial of Service (DoS) attacks, etc. It enjoys excellent data quality compared with its previous version \cite{hettich1999uci}. We follow \cite{anthi2019supervised} to use a reasonable amount of $20\%$ of the dataset during experiments. Following \cite{harb2011selecting}, we use the top-31 most informative features out of 41 features as the feature representation and denote the dataset as \textit{K}. 

\textbf{Network Intrusion Dataset: UNSW-NB15} The dataset \cite{moustafa2015unsw} was released in 2015 and was constructed on a comprehensive security testing platform commonly used by the industry. It includes normal network traffic with nine categories of modern intrusion patterns, such as DoS attack, reconnaissance attack, etc., and possesses high data quality. We perform data preprocessing to remove four features out of the original 49 features that have a value of 0 for nearly all records. We denote the dataset as \textit{N}. 

\textbf{Network Intrusion Dataset: CICIDS2017} This dataset \cite{sharafaldin2018toward} was released in 2017 and contained up-to-date intrusion trends that include seven intrusion categories, represented in 77 dimensions. We use $20\%$ of the dataset provided by its creator, and perform preprocessing steps such as categorical-numerical data conversion. We follow \cite{stiawan2020cicids} to use the top-40 most informative features, and denote the dataset as \textit{C}. 

\textbf{IoT Intrusion Dataset: UNSW-BOTIOT} This dataset \cite{koroniotis2019towards} was released in 2017. It is constructed on a realistic testbed involving commonly-used IoT devices such as the weather station, smart fridge, etc., and utilises the common lightweight IoT communication protocol MQTT. The dataset contains four up-to-date intrusion categories, represented in 46 dimensions. We follow the advice from the dataset creator to use the top-10 most informative features. The dataset is denoted as \textit{B}. 

\textbf{IoT Intrusion Dataset: UNSW-TONIOT} The dataset \cite{booij2021ton_iot} was released in 2021 and involved up-to-date IoT protocols and standards. The testbed used is sophisticated, with seven types of real IoT devices such as the GPS tracker, the weather meter, etc., and capturing heterogeneous features. The dataset contains nine types of common IoT intrusions \cite{abdelmoumin2021performance}, such as the DoS attack, scanning attack, etc. We follow \cite{qiu2020adversarial} to leverage $10\%$ of the dataset, and select two IoT devices, i.e., the GPS tracker and the weather meter, denoted as \textit{G} and \textit{W}, respectively. 

\textbf{Dataset Comprehensiveness and Intrusion Methods} The datasets used during experiments are comprehensive and representative. First, these datasets are widely recognised by the intrusion detection research community with a broad range of usage. Second, these datasets are recently released and contain modern intrusion trends and patterns, some of them are released in 2021. Third, these datasets all involve widely recognised testbeds. The IoT datasets also involve real-world IoT devices deployed in a real-world environment. Finally, the network and IoT datasets have at most eight shared intrusion categories, with a coverage of $100\%$, $55\%$, $100\%$, $100\%$ and $98\%$ on NSL-KDD, UNSW-NB15, CICIDS2017, UNSW-BOTIOT and UNSW-TONIOT, respectively. The transferrable intrusion knowledge reflect a modern intrusion trends. Hence, the datasets used are sufficient to verify the effectiveness of the OSDN model. 

\subsection{Implementation Details}\label{sec:implementation_details}

We implement the OSDN model using the deep learning framework PyTorch. The feature projectors are implemented as a single-layer neural network and use LeakyRelu as the activation function. Likewise, both the intrusion classifier $C$ and the discriminator $D$ are also implemented as single-layer neural networks. 

We apply cross validation with grid search to tune hyperparameters. Since all experiments share a single set of hyperparameter settings, the tuning effort is not too laborious. The default hyperparameter settings are as follows: $\alpha_{S}=0.8$, $\alpha_{U}=0.1$, $\beta_{S}=\beta_{T}=0.75$, $\delta=0.001$, $\theta=1.0$, $\gamma=1.0$, number of sampled dandelions $N=10$ and number of sampled unknown instances $n_{R}=100$. Additionally, the stability and robustness of the OSDN model with manipulated hyperparameters in their corresponding reasonable ranges are also verified in Section \ref{sec:hyperparameter_sensitivity_analysis}. 

During evaluation, we follow \cite{9933783} to use accuracy, category-weighted precision (P), recall (R) and F1-score (F) as evaluation metrics. Their definitions are as follows: 
\begin{equation}
  Accuracy = \frac{\sum_{k=1}^{K} (TP^{(k)} + TN^{(k)})}{n_{T}}\,,
\end{equation}
\begin{equation}
  Precision = \sum_{k=1}^{K} \frac{|\mathcal{X}_{T}^{(k)}|}{n_{T}} \cdot Precision^{(k)} = \sum_{k=1}^{K} \frac{|\mathcal{X}_{T}^{(k)}|}{n_{T}} \cdot \frac{TP^{(k)}}{TP^{(k)} + FP^{(k)}}\,,
\end{equation}
\begin{equation}
  Recall = \sum_{k=1}^{K} \frac{|\mathcal{X}_{T}^{(k)}|}{n_{T}} \cdot Recall^{(k)} = \sum_{k=1}^{K} \frac{|\mathcal{X}_{T}^{(k)}|}{n_{T}} \cdot \frac{TP^{(k)}}{TP^{(k)} + FN^{(k)}}\,,
\end{equation}
\begin{equation}
  F1 = \sum_{k=1}^{K} \frac{|\mathcal{X}_{T}^{(k)}|}{n_{T}} \cdot \frac{2 \cdot Precision^{(k)} \cdot Recall^{(k)}}{Precision^{(k)} + Recall^{(k)}}\,,
\end{equation}
where the true positive $TP^{(k)}$ denotes the number of category $k$ intrusions being correctly detected, similar for $TN^{(k)}$, $FP^{(k)}$ and $FN^{(k)}$. During experiments, we evaluate the performance in two modes: the ACC mode which evaluates the prediction with the corresponding ground truth intrusion label, and the IND mode, which treats all known and unknown intrusions as a single intrusion class. 

As an open-set DA method, following Kundu et al. \cite{kundu2020towards}, we define openness $\mathcal{O}$ as follows: 
\begin{equation}
  \mathcal{O} = 1 - \frac{K}{K'}\,.
\end{equation}
The openness $\mathcal{O}$ lies in the range between 0 and 1, the larger the openness is, the more unknown classes will be in the target II domain. 

\subsection{State-of-the-Art Baselines}\label{sec:state_of_the_art_baselines}

We use three state-of-the-art baseline methods to verify the superiority of the OSDN model, which include AMS \cite{li2022interpretable}, SR-OSDA \cite{jing2021towards} and PGL \cite{luo2020progressive}. \rw{The AMS method attempts the OSDA problem by formulating a framework with four phases. In phase 1, a discriminative representation of seen classes is learned to benefit the seen and unseen intrusion separation performed in the second phase. After performing the seen and unseen separation and the target domain is pseudo-labelled, the phase 3 further optimises the feature representation. Both phase 2 and 3 also form an iterative loop, which gradually improves the quality of intrusion recognition quality. Finally, phase 4 learns a re-projection, which promotes the generalisability of unseen intrusion recognition without sacrificing the ability to correctly recognise the seen classes. The SR-OSDA method deals with the OSDA problem by firstly separating seen and unseen intrusion instances progressively via a threshold-based pseudo-label assignment mechanism and the K-means clustering. Then, the intrusion knowledge transfer is performed by mapping both domains into a domain-invariant and discriminative feature space. Finally, the semantic information is utilised to better exploit the unknown target intrusions, so that they are not deliberately confounded together, which causes negative transfer. The PGL method integrates a graph neural network with the episodic training strategy and meanwhile applies adversarial learning to bridge the gap between two intrusion domains. In the episodic training strategy, the model progressively enlarges the labelled set via pseudo-labelling and utilise the pseudo-labelled target samples for episodic training. On top of it, the graph neural network is benefitted to perform more accurate intrusion detection. }We summarise their differences with the OSDN model as follows: 
\begin{itemize}
  \item From the dandelion-based feature space perspective, the AMS method attempts this direction. However, it lacks other mechanisms such as the graph embedding-based dandelion alignment and the dandelion compactness enhancement, and also fails to form the semantic dandelion and explore its correction effect. 
  \item From the graph embedding perspective, the PGL method utilises the graph embedding during knowledge transfer. However, the PGL completely ignores the benefit brought by utilising the graph embedding in a dandelion-based feature space. 
  \item From the semantic alignment perspective, all these methods lack effort to build a semantic hyperspherical space to guide the inter-category separation in the dandelion-based feature space, leaving a void to be filled. 
\end{itemize}
Therefore, these state-of-the-art methods are comparable and representative to verify the effectiveness of the OSDN model. 

\subsection{Intrusion Detection Performance}\label{sec:intrusion_detection_performance}

\begin{table}[!h]
  \centering
  \caption{The intrusion detection accuracy results. }
  \label{tab:giant_table}
  \begin{tabular}{c|cccccccccc}
  \Xhline{2\arrayrulewidth}
  Tasks & \multicolumn{2}{c}{K$\rightarrow$G, $\mathcal{O}$=0.6} & \multicolumn{2}{c}{N$\rightarrow$W, $\mathcal{O}$=0.4} & \multicolumn{2}{c}{C$\rightarrow$W, $\mathcal{O}$=0.5} & \multicolumn{2}{c}{K$\rightarrow$B, $\mathcal{O}$=0.5} & \multicolumn{2}{c}{K$\rightarrow$G, $\mathcal{O}$=0.2} \\ \hline
  Methods & ACC & IND & ACC & IND & ACC & IND & ACC & IND & ACC & IND \\ \hline
  AMS & 42.90 & 54.26 & 36.02 & 58.30 & 42.70 & 58.98 & 42.12 & 62.48 & 44.02 & 57.14 \\
  SROSDA & 44.02 & 57.15 & 34.32 & 57.28 & 37.25 & 57.36 & 43.13 & 62.02 & 43.56 & 55.78 \\
  PGL & 40.42 & 57.17 & 43.85 & 58.38 & 45.63 & 62.18 & 42.80 & 59.52 & 39.95 & 57.14 \\ \hline
  \rowcolor{gray}
  \textbf{OSDN} & \textbf{76.18} & \textbf{89.94} & \textbf{61.79} & \textbf{64.51} & \textbf{59.20} & \textbf{63.10} & \textbf{53.78} & \textbf{67.42} & \textbf{75.83} & \textbf{90.11} \\ \hline \hline
  Tasks & \multicolumn{2}{c}{K$\rightarrow$W, $\mathcal{O}$=0.71} & \multicolumn{2}{c}{C$\rightarrow$B, $\mathcal{O}$=0.5} & \multicolumn{2}{c}{C$\rightarrow$W, $\mathcal{O}$=0.66} & \multicolumn{2}{c}{C$\rightarrow$W, $\mathcal{O}$=0.33} & \multicolumn{2}{c}{Average} \\ \hline
  Methods & ACC & IND & ACC & IND & ACC & IND & ACC & IND & ACC & IND \\ \hline
  AMS & 38.36 & 59.44 & 49.26 & 65.92 & 41.44 & 57.98 & 42.98 & 59.22 & 42.20 & 59.30 \\
  SROSDA & 36.24 & 57.88 & 49.56 & 66.33 & 38.44 & 56.36 & 38.78 & 58.46 & 40.59 & 58.74 \\
  PGL & 34.52 & 46.89 & 51.53 & 67.68 & 39.92 & 60.10 & 45.05 & 61.41 & 42.63 & 58.94 \\ \hline
  \rowcolor{gray}
  \textbf{OSDN} & \textbf{75.05} & \textbf{78.31} & \textbf{56.22} & \textbf{69.56} & \textbf{57.32} & \textbf{62.31} & \textbf{56.39} & \textbf{64.94} & \textbf{63.53} & \textbf{72.24} \\ \Xhline{2\arrayrulewidth} 
  \end{tabular}
\end{table}

\begin{figure*}[!ht]
  \begin{center}
    \includegraphics[width=\textwidth,keepaspectratio]{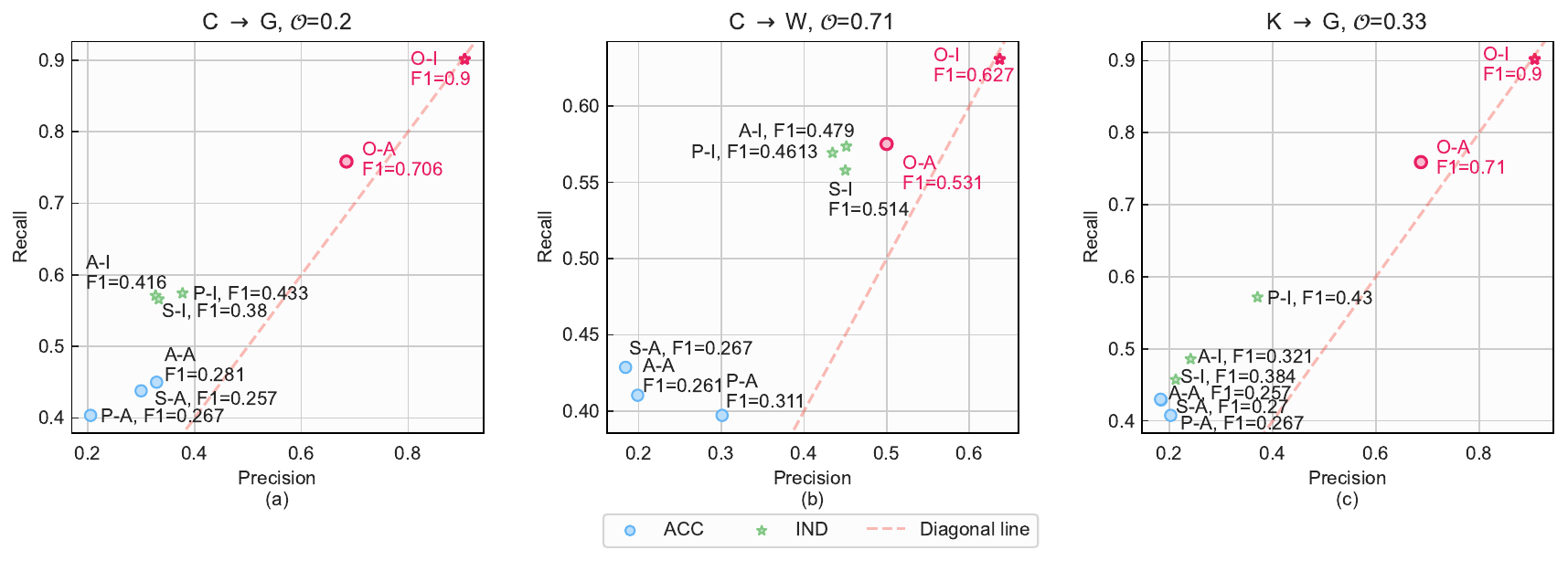}\\
    \caption{Precision, Recall and F1-Score performance on three tasks under two modes. A-A denotes the performance of method AMS under ACC mode. PGL, SR-OSDA and OSDN (Ours) are denoted as P, S and O, respectively. The X-axis and Y-axis represent precision and recall, respectively. The F1-score is marked as text in the diagram. The red diagonal line marks $f(x)=x$. }
    \vspace{-4mm}
    \label{fig:figure_precision_recall_f1score}
  \end{center}
\end{figure*}

The intrusion detection accuracy of nine randomly selected tasks with varied openness has been presented in Table \ref{tab:giant_table}. As we can observe, the OSDN model outperforms other baseline counterparts by a large margin, achieving a $20.9\%$ and $12.9\%$ performance improvement under two modes, respectively. We also measure the intrusion detection performance using three other metrics and present the results in Fig. \ref{fig:figure_precision_recall_f1score}. Under both modes, the OSDN model is positioned at the top-right corner in all three tasks, indicating that the OSDN model achieves the best precision and recall performance compared with other methods, and hence it is natural to observe the best F1-score is also yielded by the OSDN model. The best precision performance indicates the highest amount of intrusions flagged by the OSDN model are correct, while the best recall performance demonstrates the OSDN model can successfully flag as many intrusions as possible. As a harmonic mean of precision and recall, the best F1-score performance further verifies the OSDN model can elegantly balance between flagging as many intrusions as possible and simultaneously avoid triggering too many false alarms. The same result is also verified by the OSDN's nearest proximity from the red diagonal line among all methods as shown in Fig. \ref{fig:figure_precision_recall_f1score}. Hence, it demonstrates the real-world applicability of the OSDN model as an intrusion detector. 

\subsection{Robustness and Stability under Varied Openness}\label{sec:robustness_under_varied_openness}

\begin{figure*}[!ht]
  \begin{center}
    \includegraphics[width=\textwidth,keepaspectratio]{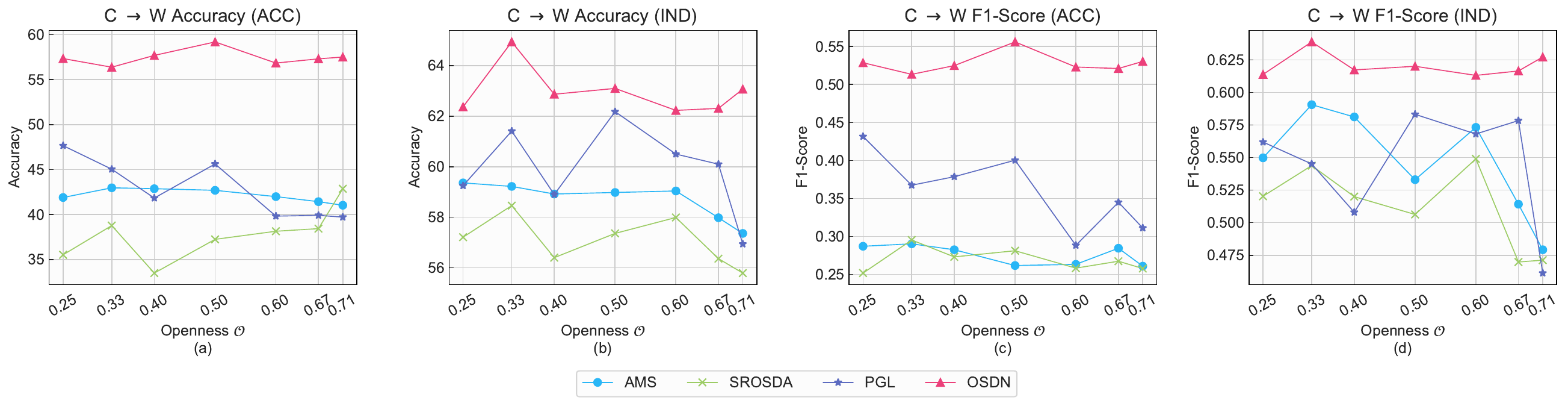}\\
    \caption{Intrusion detection accuracy and F1-score performance under varied openness levels. The accuracy results under two modes are shown in (a)-(b). The F1-score results under two modes are shown in (c)-(d). }
    \vspace{-4mm}
    \label{fig:figure_openness}
  \end{center}
\end{figure*}

\begin{figure*}[!ht]
  \begin{center}
    \includegraphics[width=\textwidth,keepaspectratio]{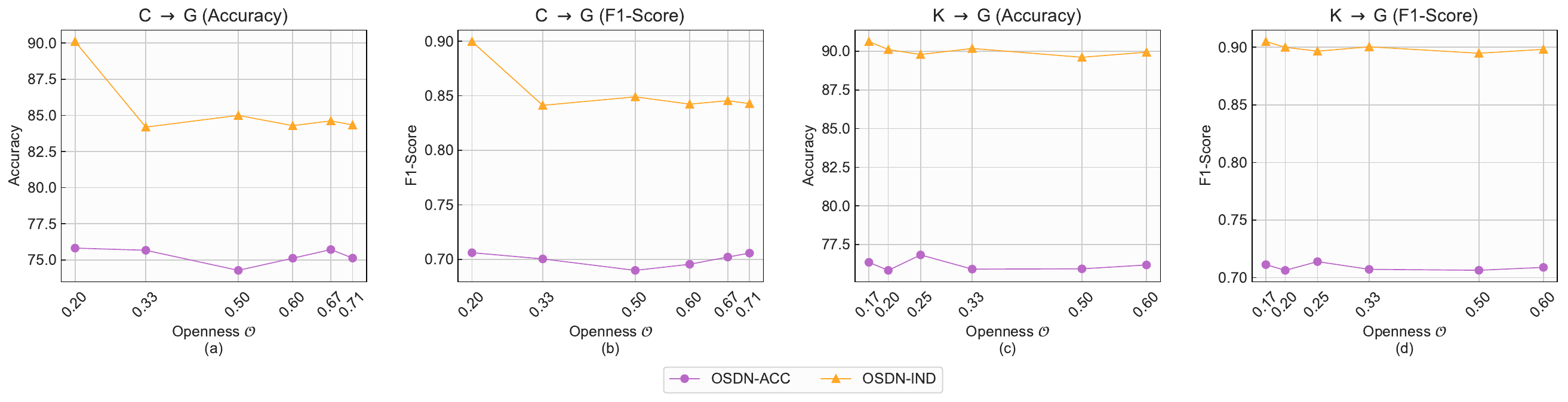}\\
    \caption{Intrusion detection accuracy and F1-score performance for two tasks under large and small openness levels in (a)-(b) and (c)-(d), respectively. }
    \vspace{-4mm}
    \label{fig:figure_stability}
  \end{center}
\end{figure*}

We first present the performance of the OSDN model and its baseline counterparts under varied openness in Fig. \ref{fig:figure_openness}. We can observe that the OSDN model stably outperforms its baseline counterparts under varied openness evaluated using both accuracy and F1-score. Besides, compared with other baseline methods, the OSDN model shows a flatter trend with less severe fluctuation. Hence, it demonstrates the robustness of the OSDN method under varied openness levels. 

We further evaluate the OSDN model against two more tasks under both large and small openness ranges. The results are shown in Fig. \ref{fig:figure_stability}. The task in Fig. \ref{fig:figure_stability} (a)-(b) has a relatively higher openness range and the task in Fig. \ref{fig:figure_stability} (c)-(d) presents a relatively lower openness range. From both Fig. \ref{fig:figure_openness} and Fig. \ref{fig:figure_stability}, the OSDN model maintains a relatively stable trend without heavy fluctuation even when the openness varied significantly. Therefore, the OSDN's capability to detect unknown intrusions in the target II domain under varied openness levels is verified and can enhance its real-world usefulness. 

\subsection{Ablation Study}\label{sec:ablation_study}

\begin{table}[!h]
  \centering
  \caption{Ablation study results for five ablation study groups. }
  \label{tab:ablation_table}
  \begin{tabular}{ccccccccccc}
  \Xhline{2\arrayrulewidth}
  \multirow{2}{*}{Group} & \multicolumn{2}{c}{\multirow{2}{*}{\begin{tabular}[c]{@{}c@{}}Experiment\\ Setting\end{tabular}}} & \multicolumn{2}{l}{N$\rightarrow$W, $\mathcal{O}$=0.40} & \multicolumn{2}{l}{C$\rightarrow$G, $\mathcal{O}$=0.71} & \multicolumn{2}{l}{K$\rightarrow$W, $\mathcal{O}$=0.71} & \multicolumn{2}{c}{Average} \\ \cline{4-11} 
   & \multicolumn{2}{c}{} & ACC & IND & ACC & IND & ACC & IND & ACC & IND \\ \hline
  A & \multicolumn{2}{c}{$\alpha_{U}=0$} & 54.87 & 60.45 & 72.36 & 82.97 & 69.00 & 73.06 & 65.41 & 72.16 \\ \hline
  B & \hspace{2mm}$\beta_{\mathcal{S}}=0$ & \hspace{2mm}$\beta_{\mathcal{T}}=0$ &  &  &  &  &  &  &  &  \\
  B1 & \hspace{2mm}\xmark & \hspace{2mm}\checkmark & 52.62 & 59.82 & 68.51 & 80.45 & 60.24 & 67.08 & 60.46 & 69.12 \\
  B2 & \hspace{2mm}\checkmark & \hspace{2mm}\xmark & 55.29 & 60.49 & 66.96 & 80.25 & 63.31 & 68.37 & 61.85 & 69.70 \\
  B3 & \hspace{2mm}\xmark & \hspace{2mm}\xmark & 54.47 & 61.44 & 72.20 & 83.72 & 67.10 & 72.23 & 64.59 & 72.46 \\ \hline
  C & \multicolumn{2}{c}{$\delta=0$} & 55.89 & 62.22 & 67.34 & 80.05 & 57.27 & 65.24 & 60.17 & 69.17 \\ \hline
  D & \multicolumn{2}{c}{$\theta=0$} & 45.15 & 58.35 & 58.10 & 77.02 & 48.22 & 58.86 & 50.49 & 64.74 \\ \hline
  \multirow{2}{*}{E} & \multicolumn{2}{c}{Discriminating} &  &  &  &  &  &  &  &  \\
   & \multicolumn{2}{c}{Strategy} &  &  &  &  &  &  &  &  \\
  E1 & \multicolumn{2}{c}{$\gamma=0$} & 56.55 & 61.61 & 67.21 & 78.43 & 59.89 & 66.71 & 61.22 & 68.92 \\
  E2 & \multicolumn{2}{c}{Domain Adv} & 54.13 & 60.03 & 72.55 & 83.69 & 60.03 & 66.70 & 62.24 & 70.14 \\ \hline
  F & \multicolumn{2}{c}{No DA} & 44.22 & 57.26 & 42.87 & 60.60 & 43.16 & 57.13 & 43.42 & 58.33 \\ \hline\hline
  \rowcolor{gray}
  \multirow{3}{*}{} & \multicolumn{2}{c}{$\alpha_{U}=0.1$,$\delta=0.001$} &  &  &  &  &  &  &  &  \\
  \rowcolor{gray}
  Full & \multicolumn{2}{c}{$\beta_{\mathcal{S}}=\beta_{\mathcal{T}}=0.75$} & \textbf{61.79} & \textbf{64.51} & \textbf{75.13} & \textbf{84.34} & \textbf{75.05} & \textbf{78.31} & \textbf{70.66} & \textbf{75.72} \\
  \rowcolor{gray}
   & \multicolumn{2}{c}{$\gamma=1.0$,$\theta=1.0$} &  &  &  &  &  &  &  &  \\ \Xhline{2\arrayrulewidth}
  \end{tabular}
\end{table}

To verify the positive contribution and the necessity of each constituting component of the OSDN model, \rw{six groups} of ablation studies are performed and the corresponding results are demonstrated in Table. \ref{tab:ablation_table}. In the ablation group \textit{A}, the unknown training mechanism is ablated, which causes the accuracy to drop around $5.3\%$ and $3.6\%$ under two modes, respectively. In the ablation group \textit{B}, either the source or target dandelion angular separation mechanism (\textit{$B_1$} and \textit{$B_2$}), or both of them (\textit{$B_3$}) are dropped. As we can observe, lacking any of the DASM will result in a significant performance reduction, hence it verifies the necessity of the DASM for both domains. Besides, using the DASM for only one domain dandelion will further deteriorate the intrusion detection efficacy. The reason is that when both DASMs are turned off, other mechanisms such as the semantic dandelion correction and discriminating sampled dandelion mechanism will still partially achieve the dandelion separation effect. However, only using a single DASM will end up with a severe dandelion misalignment. Hence, worse performance is observed for the ablation groups \textit{$B_1$} and \textit{$B_2$}. 

The dandelion embedding alignment mechanism is removed in the ablation group \textit{C}. Without it, the performance drops by $10.5\%$ and $6.6\%$ under two modes, respectively. A heavier performance drop is observed in the ablation group \textit{D}, in which the semantic dandelion correction mechanism is eliminated. Without this mechanism, there will be no semantic-assisted correction for under-separated intrusion categories, resulting in compromised intrusion detection efficacy. In the ablation group \textit{$E_1$}, the discriminating sampled dandelion mechanism is turned off, and in the ablation group \textit{$E_2$}, the traditional instance domain discriminator substitutes the proposed DSDM. As we can see, completely lacking the adversarial learning significantly hinders the intrusion detection performance, which yields a $9.4\%$ and $6.8\%$ performance reduction under two modes, respectively. Although the substituting domain adversarial learner slightly increases the performance compared with the ablation group \textit{$E_1$}, it still presents a performance that is much lower than the full OSDN model. 

\rw{Finally, in the ablation group \textit{F}, the domain adaptation mechanism is completely turned off to verify that the HDA mechanism plays an indispensable role. As we can see, removing the intrusion knowledge transfer performed by the HDA mechanism significantly degrades the intrusion detection performance, which is the worst among all ablated groups. Therefore, it justifies the necessity of having the HDA mechanism, and shows that the HDA mechanism makes a non-negligible contribution towards more accurate IoT intrusion detection. }

Overall, the full OSDN model outperforms all its ablated counterparts by a significant margin, which indicates that all constituting components of the OSDN model contribute positively towards finer intrusion knowledge transfer and hence are indispensable for achieving excellent intrusion detection performance. 

\begin{figure*}[!ht]
  \begin{center}
    \includegraphics[width=0.8\textwidth,keepaspectratio]{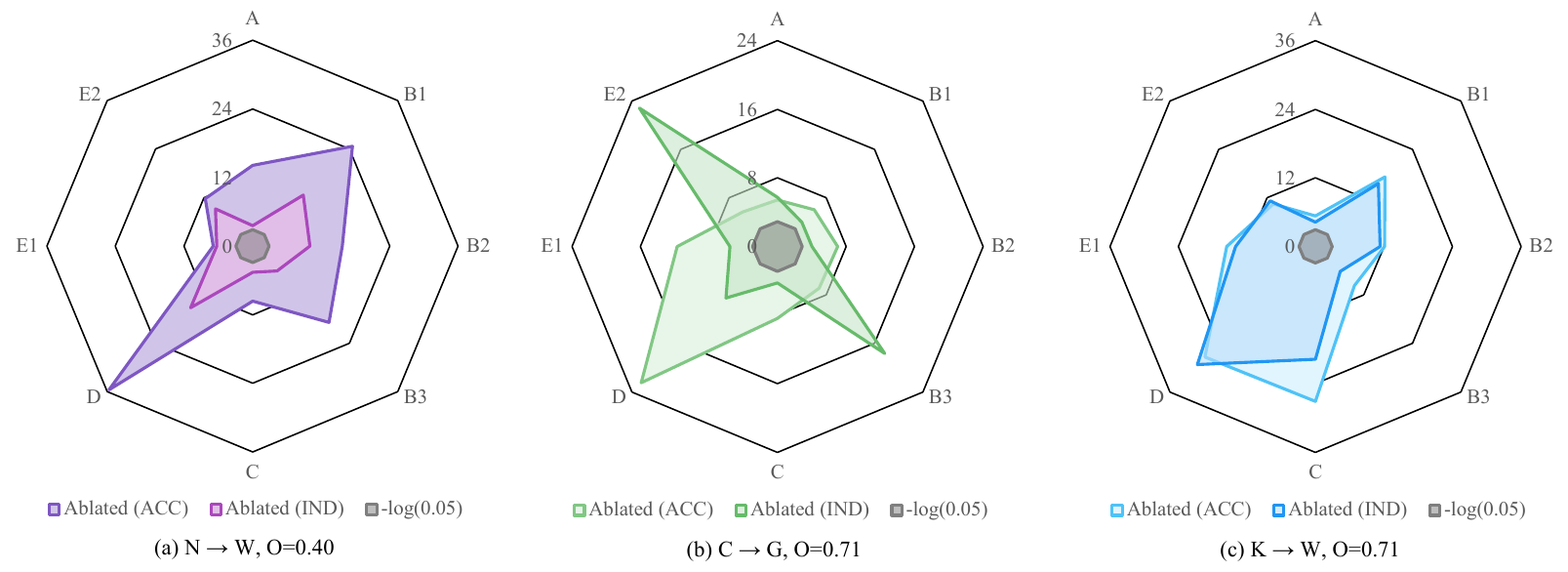}\\
    \caption{Hypothesis testing results under the significance threshold of $0.05$ to verify the statistical significance of the contribution made by each OSDN constituting component. \rw{For better visualisation, we omit ablation group F due to its significant differences compared with the full OSDN method. }}
    \vspace{-4mm}
    \label{fig:figure_hypothesis_testing}
  \end{center}
\end{figure*}

We further verify the statistical significance of each component's contribution via the significance T-test with the significance threshold of $0.05$. The results are presented in Fig. \ref{fig:figure_hypothesis_testing}. The grey area in the middle stands for the significance threshold $-log(0.05)$. Among each dimension of the radar chart, the higher the value is, the more statistically significant the contribution is for that corresponding component. As we can observe, under all three tasks and all two modes, the coloured areas present a wider coverage than the grey shaded area. The results verify the statistical soundness of all components' contributions. 

\subsection{Separability and Compactness Analysis}\label{sec:separability_and_compactness_analysis}

\begin{figure*}[!ht]
  \begin{center}
    \includegraphics[width=\textwidth,keepaspectratio]{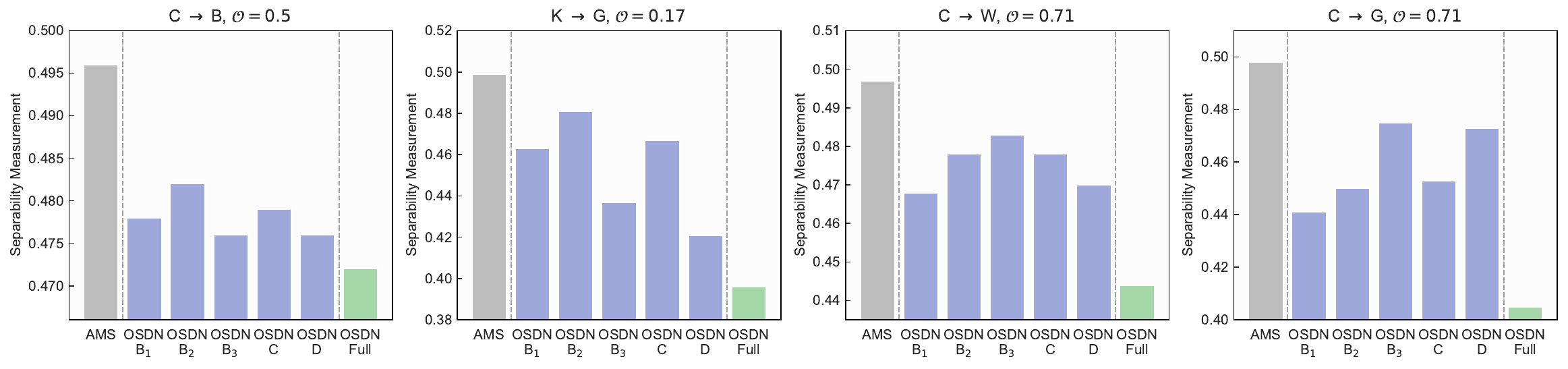}\\
    \caption{Separability measurement results on four tasks between the hyperspherical-based baseline AMS, the full OSDN model and ablated groups of the OSDN model that affect the separability. }
    \vspace{-4mm}
    \label{fig:figure_separability_measurement}
  \end{center}
\end{figure*}

The ideal common feature subspace should have the inter-category divergence as large as possible to achieve good separability and meanwhile have the intra-category variation as small as possible to achieve compactness. To verify the constituting components of the OSDN model contribute positively towards these goals, we follow Eq. \ref{equ:source_dandelion_separation_loss} to calculate the separability from an angular perspective on the source-target combined dandelion $\mathcal{DD}_{S\cup T}$, defined as follows: 
\begin{equation}
  \begin{split}
    CS_{S\cup T} &= \begin{bmatrix}
      CS_{S\cup T}^{11} & \boldsymbol{CS_{S\cup T}^{12}} & \boldsymbol{\cdots} & \boldsymbol{CS_{S\cup T}^{1K}}\\
      CS_{S\cup T}^{21} & CS_{S\cup T}^{22} & \boldsymbol{\cdots} & \boldsymbol{CS_{S\cup T}^{2K}}\\
      \vdots & \vdots & \ddots & \boldsymbol{\vdots}\\
      CS_{S\cup T}^{K1} & CS_{S\cup T}^{K2} & \cdots & CS_{S\cup T}^{KK}
    \end{bmatrix}\,,\\
    CS_{S\cup T}^{ij} &= \text{COS}(\mu_{S\cup T}^{(i)}, \mu_{S\cup T}^{(j)})\,,\\
    \mu_{S\cup T}^{(i)} &= \frac{1}{n_{S}^{(i)} + n_{T}^{(i)}} (\sum_{j=1}^{n_{S}^{(i)}} x_{S_j}^{(i)} + \sum_{j=1}^{n_{T}^{(i)}} x_{T_j}^{(i)})\,,
  \end{split}
\end{equation}
where $CS_{S\cup T}$ denotes the inter-pappus Cosine similarity matrix of the source-target combined dandelion and $CS_{S\cup T}^{ij}$ represents the Cosine similarity between the $i$\textsuperscript{th} and $j$\textsuperscript{th} pappus of the source-target combined dandelion. Then, the separability measurement $SP$ is defined as follows: 
\begin{equation}
  SP = \frac{2}{K(K-1)} \sum_{i=1}^{K-1} \sum_{j=i+1}^{K} CS_{S\cup T}^{ij}\,,
\end{equation}
the smaller the separability measurement $SP$ is, the better the separability is for the source-target combined dandelion. We present the separability measurement results between the hyperspherical-based baseline AMS, the separability-related ablated groups and the full OSDN model in Fig. \ref{fig:figure_separability_measurement}. As we observe, both the full OSDN model and its ablated groups enjoy better separability compared with the AMS baseline. Moreover, the full OSDN model presents the best inter-category separability by achieving the lowest $SP$ measurement. Hence it verifies the positive contribution of OSDN's constituting components towards enhancing inter-category separability, and the superior performance of the OSDN model over its hyperspherical-based counterpart. 

\begin{figure*}[!ht]
  \begin{center}
    \includegraphics[width=\textwidth,keepaspectratio]{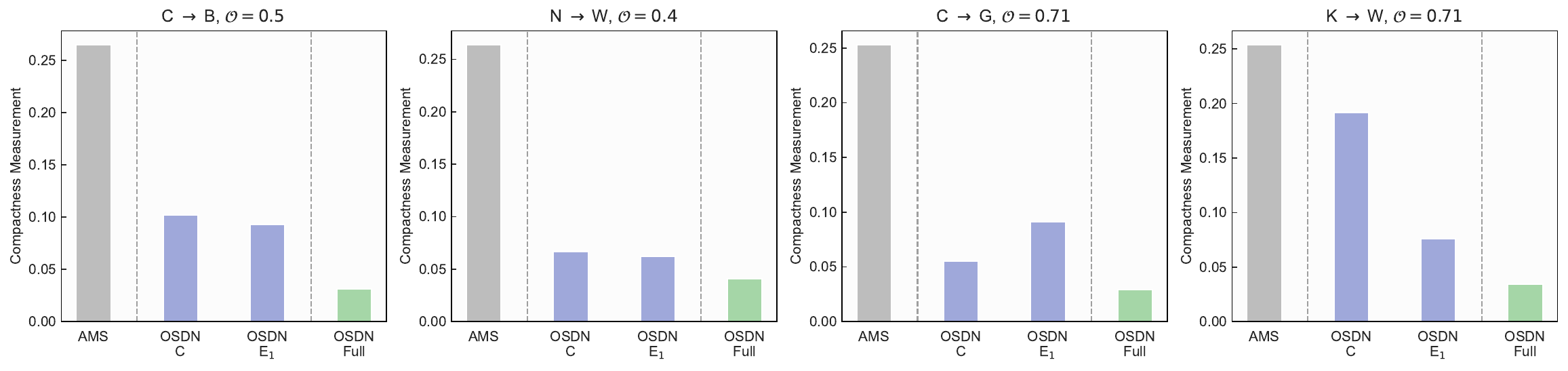}\\
    \caption{Compactness measurement results on four tasks between the hyperspherical-based baseline AMS, the full OSDN model and ablated groups of the OSDN model that affect the compactness. }
    \vspace{-4mm}
    \label{fig:figure_compactness_measurement}
  \end{center}
\end{figure*}

We then follow Eq. \ref{equ:maximum_intra_category_deviation} to measure the compactness by the average category-wise maximum deviation $d_{max}$, defined as follows: 
\begin{equation}
  d_{max} = \frac{1}{K} \sum_{i=1}^{K} d_{max}^{(i)}\,,
\end{equation}
the smaller the $d_{max}$ is, the better the compactness performance is. Again, the measurement results indicate the OSDN model outperforms both the baseline method AMS and its compactness-related ablated groups by a large margin. Hence, the excellent intra-category compactness achieved by the OSDN model is verified. 

Overall, by achieving the best inter-category separability and intra-category compactness, the OSDN model can lead to more accurate intrusion detection performance. 

\subsection{Hyperparameter Sensitivity Analysis}\label{sec:hyperparameter_sensitivity_analysis}

\begin{figure*}[!ht]
  \begin{center}
    \includegraphics[width=\textwidth,keepaspectratio]{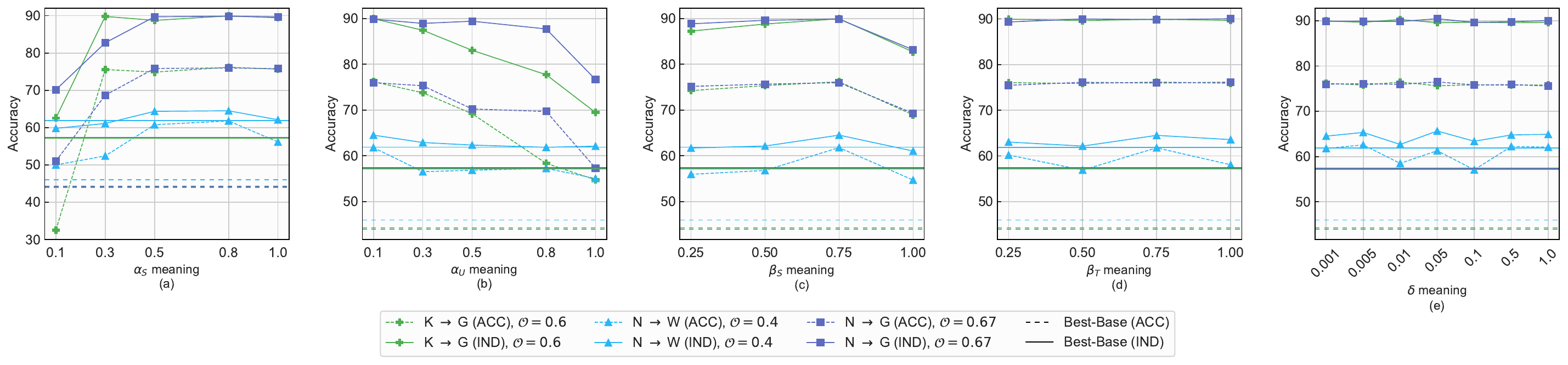}\\
    \caption{Hyperparameter sensitivity analysis for hyperparameter $\alpha_{S}$, $\alpha_{U}$, $\beta_{S}$, $\beta_{T}$ and $\delta$ under their corresponding reasonable range. The dashed lines and solid lines indicate two modes, respectively. The horizontal lines indicates the best-performed baseline counterpart. }
    \vspace{-4mm}
    \label{fig:figure_hyperparameter_sensitivity_1}
  \end{center}
\end{figure*}

\begin{figure*}[!ht]
  \begin{center}
    \includegraphics[width=\textwidth,keepaspectratio]{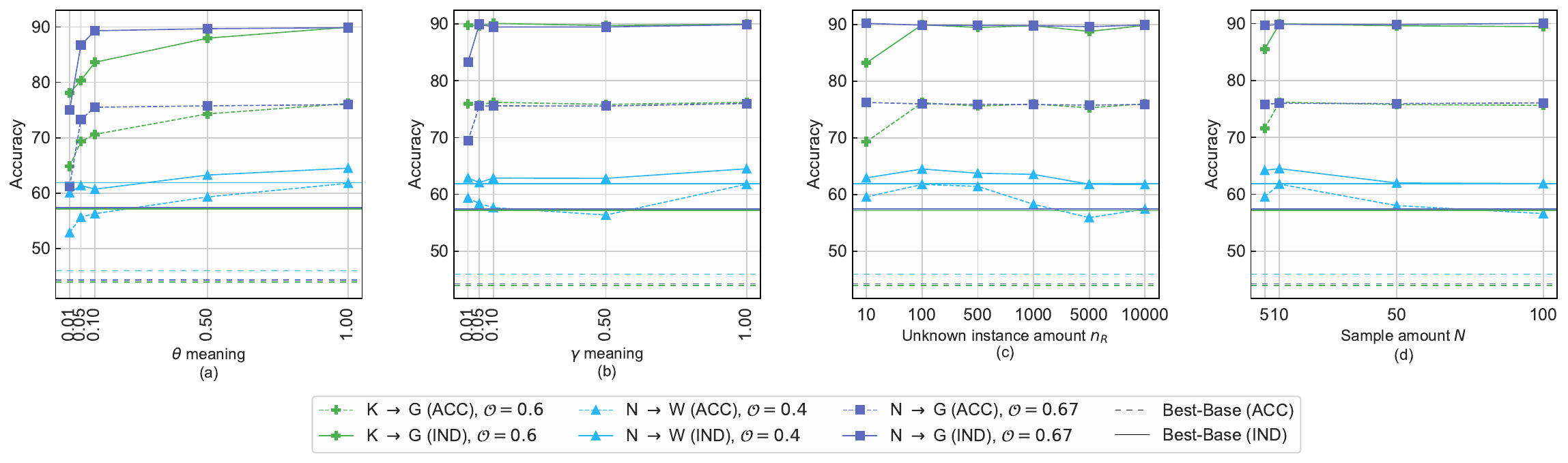}\\
    \caption{Hyperparameter sensitivity analysis for hyperparameter $\theta$, $\gamma$, unknown instance amount $n_{R}$ used in the intrusion classifier unknown training and the sampled child dandelion number $N$ used in the DSDM, under their corresponding reasonable range. }
    \vspace{-4mm}
    \label{fig:figure_hyperparameter_sensitivity_2}
  \end{center}
\end{figure*}

We verify the stability and robustness of the OSDN model under varied hyperparameter settings within their corresponding reasonable range. The results are presented in Fig. \ref{fig:figure_hyperparameter_sensitivity_1} and Fig. \ref{fig:figure_hyperparameter_sensitivity_2}. The dashed lines and solid lines indicate two modes, respectively. The horizontal lines indicates the best-performed baseline counterpart. 

We observe the OSDN model performs relatively stable without showing significant fluctuation in nearly all hyperparameter settings. As well, the OSDN model constantly outperforms the best-performed baseline method under nearly all hyperparameter settings. The OSDN model applies a single set of hyperparameter settings when facing different data domains and different tasks and still achieves such a stable level of performance. Hence, it verifies the stability and robustness of the OSDN model under manipulated hyperparameter settings. 

\subsection{Intrusion Detection Efficiency}\label{sec:intrusion_detection_efficiency}

\begin{table}[]
  \centering
  \caption{Total training time, measured in minutes. The performance of the OSDN model under different performance-sensitive hyperparameter settings are tested. }
  \label{tab:training_performance_table}
  \begin{tabular}{c|cccccccccc}
  \Xhline{2\arrayrulewidth}
  Methods & \multicolumn{3}{c}{C$\rightarrow$G,$\mathcal{O}=0.67$} & \multicolumn{3}{c}{K$\rightarrow$B,$\mathcal{O}=0.5$} & \multicolumn{3}{c}{C$\rightarrow$W,$\mathcal{O}=0.25$} & Avg \\ \hline
  AMS & \multicolumn{3}{c}{14.69} & \multicolumn{3}{c}{14.65} & \multicolumn{3}{c}{15.01} & 14.78 \\
  SROSDA & \multicolumn{3}{c}{14.57} & \multicolumn{3}{c}{14.42} & \multicolumn{3}{c}{14.55} & 14.51 \\ \hline \hline
  \rowcolor{gray}
  \multirow{3}{*}{} & \diagbox[innerwidth=0.9cm,height=0.8cm]{N}{$n_R$} & \hspace{0.3mm}=10$^2$ & \hspace{0.3mm}=10$^3$ & \diagbox[innerwidth=0.9cm,height=0.8cm]{N}{$n_R$} & \hspace{0.3mm}=10$^2$ & \hspace{0.3mm}=10$^3$ & \diagbox[innerwidth=0.9cm,height=0.8cm]{N}{$n_R$} & \hspace{0.3mm}=10$^2$ & \hspace{0.3mm}=10$^3$ &  \\
  \rowcolor{gray}
  OSDN & =10 & 4.47 & 5.46 & =10 & 4.42 & 5.41 & =10 & 4.47 & 4.76 & \textbf{9.65} \\
  \rowcolor{gray}
   & =10$^2$ & 14.06 & 15.00 & =10$^2$ & 14.49 & 14.43 & =10$^2$ & 14.02 & 14.83 &  \\ \Xhline{2\arrayrulewidth}
  \end{tabular}
\end{table}

\begin{table}[]
  \centering
  \caption{Inference time per network traffic instance, measured in milliseconds ($10^{-3}$ second). The performance of the OSDN model under different performance-sensitive hyperparameter settings are tested. }
  \label{tab:inference_performance_table}
  \begin{tabular}{c|cccccccccc}
  \Xhline{2\arrayrulewidth}
  Methods & \multicolumn{3}{c}{C$\rightarrow$G,$\mathcal{O}=0.67$} & \multicolumn{3}{c}{K$\rightarrow$B,$\mathcal{O}=0.5$} & \multicolumn{3}{c}{C$\rightarrow$W,$\mathcal{O}=0.25$} & Avg \\ \hline
  AMS & \multicolumn{3}{c}{1.68} & \multicolumn{3}{c}{1.65} & \multicolumn{3}{c}{1.68} & 1.67 \\
  SROSDA & \multicolumn{3}{c}{1.63} & \multicolumn{3}{c}{1.63} & \multicolumn{3}{c}{1.67} & 1.64 \\ \hline \hline
  \rowcolor{gray}
  \multirow{3}{*}{} & \diagbox[innerwidth=0.9cm,height=0.8cm]{N}{$n_R$} & \hspace{0.3mm}=10$^2$ & \hspace{0.3mm}=10$^3$ & \diagbox[innerwidth=0.9cm,height=0.8cm]{N}{$n_R$} & \hspace{0.3mm}=10$^2$ & \hspace{0.3mm}=10$^3$ & \diagbox[innerwidth=0.9cm,height=0.8cm]{N}{$n_R$} & \hspace{0.3mm}=10$^2$ & \hspace{0.3mm}=10$^3$ &  \\
  \rowcolor{gray}
  OSDN & =10 & 0.100 & 0.099 & =10 & 0.098 & 0.099 & =10 & 0.101 & 0.103 & \textbf{0.100} \\
  \rowcolor{gray}
   & =10$^2$ & 0.098 & 0.100 & =10$^2$ & 0.099 & 0.100 & =10$^2$ & 0.101 & 0.102 &  \\ \Xhline{2\arrayrulewidth}
  \end{tabular}
\end{table}

We finally verify the training and intrusion detection inference efficiency of the OSDN model. The training time taken has been summarised in Table \ref{tab:training_performance_table}, and the inference time per network traffic instance has been summarised in Table \ref{tab:inference_performance_table}. We only compare the OSDN model with the top-two best-performing baseline methods. As shown in Table \ref{tab:training_performance_table}, under varied settings of the OSDN model, the OSDN model performs more efficiently compared with its counterparts in nearly all settings. Since the model training can be performed on computationally-sufficient devices such as network gateway servers, therefore, the training efficiency of the OSDN model is satisfactory. Besides, as indicated in Table \ref{tab:inference_performance_table}, the OSDN model significantly outperforms its baseline counterparts in terms of the inference time taken to examine a network traffic instance. Therefore, the results verify the efficiency of the OSDN model, and demonstrate its real-world applicability as an efficient and accurate intrusion detector. 

\section{Conclusion}\label{sec:conclusion}

In this paper, we propose the Open-Set Dandelion Network (OSDN) based on unsupervised heterogeneous domain adaptation in an open-set manner. The OSDN model tackles the IoT data scarcity by transferring intrusion knowledge from source NI domain to promote more accurate intrusion detection for the target IoT domain. The relaxation of the closed-set assumption lets the OSDN model detect both known and newly-emerged unknown intrusions in the IoT intrusion domain, hence it is more applicable in the real-world. The OSDN model achieves this by first forming the source domain into a dandelion-like feature space that emphasises inter-category separability and intra-category compactness. Then, the dandelion-based target membership mechanism constructs the target dandelion for intrusion knowledge transfer. The dandelion angular separation mechanism is used to promote inter-category separability, while the dandelion embedding alignment mechanism facilitates knowledge transfer from a graph embedding perspective. Also, the discriminating sampled dandelion mechanism is used to promote intra-category compactness. Trained using both known and generated unknown intrusion information, the intrusion classifier yields probabilistic semantics that can emphasise easily-confused categories and hence provide correction for the inter-category separation mechanism. Holistically, these mechanisms form the OSDN model and benefit in a more effective intrusion detection for IoT scenarios. Comprehensive experiments on five intrusion datasets are conducted. The OSDN model outperforms three state-of-the-art baseline methods by $16.9\%$. The effectiveness of each OSDN constituting component, the stability and the efficiency of the OSDN model are also verified. \rw{For future research, it is worthwhile to extend the OSDN model to the multi-source setting, in which the intrusion knowledge from multiple source domains can jointly benefit the open-set intrusion knowledge transfer. Besides, we can also consider using a category-wise attention mechanism during the intrusion knowledge transfer to account for diverse knowledge transfer sufficiency for each intrusion category. We will leave these as our future research directions. }

\begin{acks}
We would like to express out sincere gratitude to Prof. André Brinkmann for his invaluable assistance in enhancing the quality of this paper. This work is supported by the National Key R\&D Program of China (No. 2021YFB3300200), National Natural Science Foundation of China (No. 92267105), Guangdong Special Support Plan (No. 2021TQ06X990), the Third Xinjiang Scientific Expedition Program (Grant No. 2021XJKK1300), Shenzhen Science and Technology Plan Project (Shenzhen-Hong Kong-Macau Category C, No. SGDX20220530111001003), Key-Area Research and Development Program of Guangdong Province (No. 2021B010140005) and the Chinese Academy of Sciences President's International Fellowship Initiative (Grant No. 2023VTA0001, No. 2023DT0003). 
\end{acks}

\bibliographystyle{ACM-Reference-Format}
\bibliography{OSDN}


\begin{thebibliography}{44}


\ifx \showCODEN    \undefined \def \showCODEN     #1{\unskip}     \fi
\ifx \showDOI      \undefined \def \showDOI       #1{#1}\fi
\ifx \showISBNx    \undefined \def \showISBNx     #1{\unskip}     \fi
\ifx \showISBNxiii \undefined \def \showISBNxiii  #1{\unskip}     \fi
\ifx \showISSN     \undefined \def \showISSN      #1{\unskip}     \fi
\ifx \showLCCN     \undefined \def \showLCCN      #1{\unskip}     \fi
\ifx \shownote     \undefined \def \shownote      #1{#1}          \fi
\ifx \showarticletitle \undefined \def \showarticletitle #1{#1}   \fi
\ifx \showURL      \undefined \def \showURL       {\relax}        \fi
\providecommand\bibfield[2]{#2}
\providecommand\bibinfo[2]{#2}
\providecommand\natexlab[1]{#1}
\providecommand\showeprint[2][]{arXiv:#2}

\bibitem[Abdelmoumin et~al\mbox{.}(2021)]%
        {abdelmoumin2021performance}
\bibfield{author}{\bibinfo{person}{Ghada Abdelmoumin}, \bibinfo{person}{Danda~B
  Rawat}, {and} \bibinfo{person}{Abdul Rahman}.}
  \bibinfo{year}{2021}\natexlab{}.
\newblock \showarticletitle{On the performance of machine learning models for
  anomaly-based intelligent intrusion detection systems for the internet of
  things}.
\newblock \bibinfo{journal}{\emph{IEEE Internet of Things Journal}}
  \bibinfo{volume}{9}, \bibinfo{number}{6} (\bibinfo{year}{2021}),
  \bibinfo{pages}{4280--4290}.
\newblock


\bibitem[Anthi et~al\mbox{.}(2019)]%
        {anthi2019supervised}
\bibfield{author}{\bibinfo{person}{Eirini Anthi}, \bibinfo{person}{Lowri
  Williams}, \bibinfo{person}{Ma{\l}gorzata S{\l}owi{\'n}ska},
  \bibinfo{person}{George Theodorakopoulos}, {and} \bibinfo{person}{Pete
  Burnap}.} \bibinfo{year}{2019}\natexlab{}.
\newblock \showarticletitle{A supervised intrusion detection system for smart
  home IoT devices}.
\newblock \bibinfo{journal}{\emph{IEEE Internet of Things Journal}}
  \bibinfo{volume}{6}, \bibinfo{number}{5} (\bibinfo{year}{2019}),
  \bibinfo{pages}{9042--9053}.
\newblock


\bibitem[Booij et~al\mbox{.}(2021)]%
        {booij2021ton_iot}
\bibfield{author}{\bibinfo{person}{Tim~M Booij}, \bibinfo{person}{Irina
  Chiscop}, \bibinfo{person}{Erik Meeuwissen}, \bibinfo{person}{Nour Moustafa},
  {and} \bibinfo{person}{Frank~TH den Hartog}.}
  \bibinfo{year}{2021}\natexlab{}.
\newblock \showarticletitle{ToN\_IoT: The role of heterogeneity and the need
  for standardization of features and attack types in IoT network intrusion
  data sets}.
\newblock \bibinfo{journal}{\emph{IEEE Internet of Things Journal}}
  \bibinfo{volume}{9}, \bibinfo{number}{1} (\bibinfo{year}{2021}),
  \bibinfo{pages}{485--496}.
\newblock


\bibitem[Bovenzi et~al\mbox{.}(2023)]%
        {bovenzi2023network}
\bibfield{author}{\bibinfo{person}{Giampaolo Bovenzi},
  \bibinfo{person}{Giuseppe Aceto}, \bibinfo{person}{Domenico Ciuonzo},
  \bibinfo{person}{Antonio Montieri}, \bibinfo{person}{Valerio Persico}, {and}
  \bibinfo{person}{Antonio Pescap{\'e}}.} \bibinfo{year}{2023}\natexlab{}.
\newblock \showarticletitle{Network anomaly detection methods in IoT
  environments via deep learning: A Fair comparison of performance and
  robustness}.
\newblock \bibinfo{journal}{\emph{Computers \& Security}}
  \bibinfo{volume}{128} (\bibinfo{year}{2023}), \bibinfo{pages}{103167}.
\newblock


\bibitem[Bovenzi et~al\mbox{.}(2020)]%
        {bovenzi2020hierarchical}
\bibfield{author}{\bibinfo{person}{Giampaolo Bovenzi},
  \bibinfo{person}{Giuseppe Aceto}, \bibinfo{person}{Domenico Ciuonzo},
  \bibinfo{person}{Valerio Persico}, {and} \bibinfo{person}{Antonio
  Pescap{\'e}}.} \bibinfo{year}{2020}\natexlab{}.
\newblock \showarticletitle{A hierarchical hybrid intrusion detection approach
  in IoT scenarios}. In \bibinfo{booktitle}{\emph{GLOBECOM 2020-2020 IEEE
  global communications conference}}. IEEE, \bibinfo{pages}{1--7}.
\newblock


\bibitem[Dai et~al\mbox{.}(2023)]%
        {10049541}
\bibfield{author}{\bibinfo{person}{Hao Dai}, \bibinfo{person}{Jiashu Wu},
  \bibinfo{person}{Yang Wang}, \bibinfo{person}{Jerome Yen},
  \bibinfo{person}{Yong Zhang}, {and} \bibinfo{person}{Chengzhong Xu}.}
  \bibinfo{year}{2023}\natexlab{}.
\newblock \showarticletitle{Cost-Efficient Sharing Algorithms for DNN Model
  Serving in Mobile Edge Networks}.
\newblock \bibinfo{journal}{\emph{IEEE Transactions on Services Computing}}
  (\bibinfo{year}{2023}), \bibinfo{pages}{1--14}.
\newblock
\urldef\tempurl%
\url{https://doi.org/10.1109/TSC.2023.3247049}
\showDOI{\tempurl}


\bibitem[Dietz et~al\mbox{.}(2018)]%
        {dietz2018iot}
\bibfield{author}{\bibinfo{person}{Christian Dietz},
  \bibinfo{person}{Raphael~Labaca Castro}, \bibinfo{person}{Jessica
  Steinberger}, \bibinfo{person}{Cezary Wilczak}, \bibinfo{person}{Marcel
  Antzek}, \bibinfo{person}{Anna Sperotto}, {and} \bibinfo{person}{Aiko Pras}.}
  \bibinfo{year}{2018}\natexlab{}.
\newblock \showarticletitle{IoT-botnet detection and isolation by access
  routers}. In \bibinfo{booktitle}{\emph{2018 9th International Conference on
  the Network of the Future (NOF)}}. IEEE, \bibinfo{pages}{88--95}.
\newblock


\bibitem[Erlacher and Dressler(2022)]%
        {8999496}
\bibfield{author}{\bibinfo{person}{Felix Erlacher} {and} \bibinfo{person}{Falko
  Dressler}.} \bibinfo{year}{2022}\natexlab{}.
\newblock \showarticletitle{On High-Speed Flow-Based Intrusion Detection Using
  Snort-Compatible Signatures}.
\newblock \bibinfo{journal}{\emph{IEEE Transactions on Dependable and Secure
  Computing}} \bibinfo{volume}{19}, \bibinfo{number}{1} (\bibinfo{year}{2022}),
  \bibinfo{pages}{495--506}.
\newblock
\urldef\tempurl%
\url{https://doi.org/10.1109/TDSC.2020.2973992}
\showDOI{\tempurl}


\bibitem[Eskandari et~al\mbox{.}(2020)]%
        {eskandari2020passban}
\bibfield{author}{\bibinfo{person}{Mojtaba Eskandari},
  \bibinfo{person}{Zaffar~Haider Janjua}, \bibinfo{person}{Massimo Vecchio},
  {and} \bibinfo{person}{Fabio Antonelli}.} \bibinfo{year}{2020}\natexlab{}.
\newblock \showarticletitle{Passban IDS: An intelligent anomaly-based intrusion
  detection system for IoT edge devices}.
\newblock \bibinfo{journal}{\emph{IEEE Internet of Things Journal}}
  \bibinfo{volume}{7}, \bibinfo{number}{8} (\bibinfo{year}{2020}),
  \bibinfo{pages}{6882--6897}.
\newblock


\bibitem[Fang et~al\mbox{.}(2020)]%
        {fang2020open}
\bibfield{author}{\bibinfo{person}{Zhen Fang}, \bibinfo{person}{Jie Lu},
  \bibinfo{person}{Feng Liu}, \bibinfo{person}{Junyu Xuan}, {and}
  \bibinfo{person}{Guangquan Zhang}.} \bibinfo{year}{2020}\natexlab{}.
\newblock \showarticletitle{Open set domain adaptation: Theoretical bound and
  algorithm}.
\newblock \bibinfo{journal}{\emph{IEEE transactions on neural networks and
  learning systems}} \bibinfo{volume}{32}, \bibinfo{number}{10}
  (\bibinfo{year}{2020}), \bibinfo{pages}{4309--4322}.
\newblock


\bibitem[Ganin et~al\mbox{.}(2016)]%
        {ganin2016domain}
\bibfield{author}{\bibinfo{person}{Yaroslav Ganin}, \bibinfo{person}{Evgeniya
  Ustinova}, \bibinfo{person}{Hana Ajakan}, \bibinfo{person}{Pascal Germain},
  \bibinfo{person}{Hugo Larochelle}, \bibinfo{person}{Fran{\c{c}}ois
  Laviolette}, \bibinfo{person}{Mario Marchand}, {and} \bibinfo{person}{Victor
  Lempitsky}.} \bibinfo{year}{2016}\natexlab{}.
\newblock \showarticletitle{Domain-adversarial training of neural networks}.
\newblock \bibinfo{journal}{\emph{The journal of machine learning research}}
  \bibinfo{volume}{17}, \bibinfo{number}{1} (\bibinfo{year}{2016}),
  \bibinfo{pages}{2096--2030}.
\newblock


\bibitem[Harb et~al\mbox{.}(2011)]%
        {harb2011selecting}
\bibfield{author}{\bibinfo{person}{Hany~M Harb}, \bibinfo{person}{Afaf~A
  Zaghrot}, \bibinfo{person}{Mohamed~A Gomaa}, {and} \bibinfo{person}{Abeer~S
  Desuky}.} \bibinfo{year}{2011}\natexlab{}.
\newblock \showarticletitle{Selecting optimal subset of features for intrusion
  detection systems}.
\newblock \bibinfo{journal}{\emph{Advances in Computational Sciences and
  Technology}} \bibinfo{volume}{4}, \bibinfo{number}{2} (\bibinfo{year}{2011}),
  \bibinfo{pages}{179--192}.
\newblock


\bibitem[Hettich(1999)]%
        {hettich1999uci}
\bibfield{author}{\bibinfo{person}{Setz Hettich}.}
  \bibinfo{year}{1999}\natexlab{}.
\newblock \showarticletitle{The uci kdd archive}.
\newblock \bibinfo{journal}{\emph{http://kdd. ics. uci. edu}}
  (\bibinfo{year}{1999}).
\newblock


\bibitem[Jing et~al\mbox{.}(2021)]%
        {jing2021towards}
\bibfield{author}{\bibinfo{person}{Taotao Jing}, \bibinfo{person}{Hongfu Liu},
  {and} \bibinfo{person}{Zhengming Ding}.} \bibinfo{year}{2021}\natexlab{}.
\newblock \showarticletitle{Towards novel target discovery through open-set
  domain adaptation}. In \bibinfo{booktitle}{\emph{Proceedings of the IEEE/CVF
  International Conference on Computer Vision}}. \bibinfo{pages}{9322--9331}.
\newblock


\bibitem[Koroniotis et~al\mbox{.}(2019)]%
        {koroniotis2019towards}
\bibfield{author}{\bibinfo{person}{Nickolaos Koroniotis}, \bibinfo{person}{Nour
  Moustafa}, \bibinfo{person}{Elena Sitnikova}, {and} \bibinfo{person}{Benjamin
  Turnbull}.} \bibinfo{year}{2019}\natexlab{}.
\newblock \showarticletitle{Towards the development of realistic botnet dataset
  in the internet of things for network forensic analytics: Bot-iot dataset}.
\newblock \bibinfo{journal}{\emph{Future Generation Computer Systems}}
  \bibinfo{volume}{100} (\bibinfo{year}{2019}), \bibinfo{pages}{779--796}.
\newblock


\bibitem[Kundu et~al\mbox{.}(2020)]%
        {kundu2020towards}
\bibfield{author}{\bibinfo{person}{Jogendra~Nath Kundu},
  \bibinfo{person}{Naveen Venkat}, \bibinfo{person}{Ambareesh Revanur},
  \bibinfo{person}{R~Venkatesh Babu}, {et~al\mbox{.}}}
  \bibinfo{year}{2020}\natexlab{}.
\newblock \showarticletitle{Towards inheritable models for open-set domain
  adaptation}. In \bibinfo{booktitle}{\emph{Proceedings of the IEEE/CVF
  conference on computer vision and pattern recognition}}.
  \bibinfo{pages}{12376--12385}.
\newblock


\bibitem[Li et~al\mbox{.}(2020)]%
        {10.1145/3394171.3413995}
\bibfield{author}{\bibinfo{person}{Shuang Li}, \bibinfo{person}{Binhui Xie},
  \bibinfo{person}{Jiashu Wu}, \bibinfo{person}{Ying Zhao},
  \bibinfo{person}{Chi~Harold Liu}, {and} \bibinfo{person}{Zhengming Ding}.}
  \bibinfo{year}{2020}\natexlab{}.
\newblock \bibinfo{booktitle}{\emph{Simultaneous Semantic Alignment Network for
  Heterogeneous Domain Adaptation}}.
\newblock \bibinfo{publisher}{Association for Computing Machinery},
  \bibinfo{address}{New York, NY, USA}, \bibinfo{pages}{3866–3874}.
\newblock
\showISBNx{9781450379885}
\urldef\tempurl%
\url{https://doi.org/10.1145/3394171.3413995}
\showURL{%
\tempurl}


\bibitem[Li et~al\mbox{.}(2022)]%
        {li2022interpretable}
\bibfield{author}{\bibinfo{person}{Xinhao Li}, \bibinfo{person}{Jingjing Li},
  \bibinfo{person}{Zhekai Du}, \bibinfo{person}{Lei Zhu}, {and}
  \bibinfo{person}{Wen Li}.} \bibinfo{year}{2022}\natexlab{}.
\newblock \showarticletitle{Interpretable Open-Set Domain Adaptation via
  Angular Margin Separation}. In \bibinfo{booktitle}{\emph{Computer
  Vision--ECCV 2022: 17th European Conference, Tel Aviv, Israel, October
  23--27, 2022, Proceedings, Part XXXIV}}. Springer, \bibinfo{pages}{1--18}.
\newblock


\bibitem[Liang et~al\mbox{.}(2021)]%
        {liang2021pareto}
\bibfield{author}{\bibinfo{person}{Jian Liang}, \bibinfo{person}{Kaixiong
  Gong}, \bibinfo{person}{Shuang Li}, \bibinfo{person}{Chi~Harold Liu},
  \bibinfo{person}{Han Li}, \bibinfo{person}{Di Liu}, \bibinfo{person}{Guoren
  Wang}, {et~al\mbox{.}}} \bibinfo{year}{2021}\natexlab{}.
\newblock \showarticletitle{Pareto domain adaptation}.
\newblock \bibinfo{journal}{\emph{Advances in Neural Information Processing
  Systems}}  \bibinfo{volume}{34} (\bibinfo{year}{2021}),
  \bibinfo{pages}{12917--12929}.
\newblock


\bibitem[Lu and Da~Xu(2018)]%
        {lu2018internet}
\bibfield{author}{\bibinfo{person}{Yang Lu} {and} \bibinfo{person}{Li Da~Xu}.}
  \bibinfo{year}{2018}\natexlab{}.
\newblock \showarticletitle{Internet of Things (IoT) cybersecurity research: A
  review of current research topics}.
\newblock \bibinfo{journal}{\emph{IEEE Internet of Things Journal}}
  \bibinfo{volume}{6}, \bibinfo{number}{2} (\bibinfo{year}{2018}),
  \bibinfo{pages}{2103--2115}.
\newblock


\bibitem[Luo et~al\mbox{.}(2020)]%
        {luo2020progressive}
\bibfield{author}{\bibinfo{person}{Yadan Luo}, \bibinfo{person}{Zijian Wang},
  \bibinfo{person}{Zi Huang}, {and} \bibinfo{person}{Mahsa Baktashmotlagh}.}
  \bibinfo{year}{2020}\natexlab{}.
\newblock \showarticletitle{Progressive graph learning for open-set domain
  adaptation}. In \bibinfo{booktitle}{\emph{International Conference on Machine
  Learning}}. PMLR, \bibinfo{pages}{6468--6478}.
\newblock


\bibitem[Mangino et~al\mbox{.}(2020)]%
        {10.1145/3394504}
\bibfield{author}{\bibinfo{person}{Antonio Mangino},
  \bibinfo{person}{Morteza~Safaei Pour}, {and} \bibinfo{person}{Elias
  Bou-Harb}.} \bibinfo{year}{2020}\natexlab{}.
\newblock \showarticletitle{Internet-Scale Insecurity of Consumer Internet of
  Things: An Empirical Measurements Perspective}.
\newblock \bibinfo{journal}{\emph{ACM Trans. Manage. Inf. Syst.}}
  \bibinfo{volume}{11}, \bibinfo{number}{4}, Article \bibinfo{articleno}{21}
  (\bibinfo{date}{oct} \bibinfo{year}{2020}), \bibinfo{numpages}{24}~pages.
\newblock
\showISSN{2158-656X}
\urldef\tempurl%
\url{https://doi.org/10.1145/3394504}
\showDOI{\tempurl}


\bibitem[Mehedi et~al\mbox{.}(2022)]%
        {mehedi2022dependable}
\bibfield{author}{\bibinfo{person}{Sk~Tanzir Mehedi}, \bibinfo{person}{Adnan
  Anwar}, \bibinfo{person}{Ziaur Rahman}, \bibinfo{person}{Kawsar Ahmed}, {and}
  \bibinfo{person}{Rafiqul Islam}.} \bibinfo{year}{2022}\natexlab{}.
\newblock \showarticletitle{Dependable intrusion detection system for IoT: A
  deep transfer learning based approach}.
\newblock \bibinfo{journal}{\emph{IEEE Transactions on Industrial Informatics}}
  \bibinfo{volume}{19}, \bibinfo{number}{1} (\bibinfo{year}{2022}),
  \bibinfo{pages}{1006--1017}.
\newblock


\bibitem[Mirsky et~al\mbox{.}(2018)]%
        {mirsky2018kitsune}
\bibfield{author}{\bibinfo{person}{Yisroel Mirsky}, \bibinfo{person}{Tomer
  Doitshman}, \bibinfo{person}{Yuval Elovici}, {and} \bibinfo{person}{Asaf
  Shabtai}.} \bibinfo{year}{2018}\natexlab{}.
\newblock \showarticletitle{Kitsune: an ensemble of autoencoders for online
  network intrusion detection}.
\newblock \bibinfo{journal}{\emph{arXiv preprint arXiv:1802.09089}}
  (\bibinfo{year}{2018}).
\newblock


\bibitem[Mitchell and Chen(2013a)]%
        {mitchell2013adaptive}
\bibfield{author}{\bibinfo{person}{Robert Mitchell} {and} \bibinfo{person}{Ray
  Chen}.} \bibinfo{year}{2013}\natexlab{a}.
\newblock \showarticletitle{Adaptive intrusion detection of malicious unmanned
  air vehicles using behavior rule specifications}.
\newblock \bibinfo{journal}{\emph{IEEE transactions on systems, man, and
  cybernetics: systems}} \bibinfo{volume}{44}, \bibinfo{number}{5}
  (\bibinfo{year}{2013}), \bibinfo{pages}{593--604}.
\newblock


\bibitem[Mitchell and Chen(2013b)]%
        {mitchell2013behavior}
\bibfield{author}{\bibinfo{person}{Robert Mitchell} {and} \bibinfo{person}{Ray
  Chen}.} \bibinfo{year}{2013}\natexlab{b}.
\newblock \showarticletitle{Behavior-rule based intrusion detection systems for
  safety critical smart grid applications}.
\newblock \bibinfo{journal}{\emph{IEEE Transactions on Smart Grid}}
  \bibinfo{volume}{4}, \bibinfo{number}{3} (\bibinfo{year}{2013}),
  \bibinfo{pages}{1254--1263}.
\newblock


\bibitem[Moustafa and Slay(2015)]%
        {moustafa2015unsw}
\bibfield{author}{\bibinfo{person}{Nour Moustafa} {and} \bibinfo{person}{Jill
  Slay}.} \bibinfo{year}{2015}\natexlab{}.
\newblock \showarticletitle{UNSW-NB15: a comprehensive data set for network
  intrusion detection systems (UNSW-NB15 network data set)}. In
  \bibinfo{booktitle}{\emph{2015 military communications and information
  systems conference (MilCIS)}}. IEEE, \bibinfo{pages}{1--6}.
\newblock


\bibitem[Muhammad et~al\mbox{.}(2020)]%
        {muhammad2020stacked}
\bibfield{author}{\bibinfo{person}{Ghulam Muhammad}, \bibinfo{person}{M~Shamim
  Hossain}, {and} \bibinfo{person}{Sahil Garg}.}
  \bibinfo{year}{2020}\natexlab{}.
\newblock \showarticletitle{Stacked autoencoder-based intrusion detection
  system to combat financial fraudulent}.
\newblock \bibinfo{journal}{\emph{IEEE Internet of Things Journal}}
  (\bibinfo{year}{2020}).
\newblock


\bibitem[Murali and Jamalipour(2019)]%
        {murali2019lightweight}
\bibfield{author}{\bibinfo{person}{Sarumathi Murali} {and}
  \bibinfo{person}{Abbas Jamalipour}.} \bibinfo{year}{2019}\natexlab{}.
\newblock \showarticletitle{A lightweight intrusion detection for sybil attack
  under mobile RPL in the internet of things}.
\newblock \bibinfo{journal}{\emph{IEEE Internet of Things Journal}}
  \bibinfo{volume}{7}, \bibinfo{number}{1} (\bibinfo{year}{2019}),
  \bibinfo{pages}{379--388}.
\newblock


\bibitem[Panareda~Busto and Gall(2017)]%
        {panareda2017open}
\bibfield{author}{\bibinfo{person}{Pau Panareda~Busto} {and}
  \bibinfo{person}{Juergen Gall}.} \bibinfo{year}{2017}\natexlab{}.
\newblock \showarticletitle{Open set domain adaptation}. In
  \bibinfo{booktitle}{\emph{Proceedings of the IEEE international conference on
  computer vision}}. \bibinfo{pages}{754--763}.
\newblock


\bibitem[Qiu et~al\mbox{.}(2020)]%
        {qiu2020adversarial}
\bibfield{author}{\bibinfo{person}{Han Qiu}, \bibinfo{person}{Tian Dong},
  \bibinfo{person}{Tianwei Zhang}, \bibinfo{person}{Jialiang Lu},
  \bibinfo{person}{Gerard Memmi}, {and} \bibinfo{person}{Meikang Qiu}.}
  \bibinfo{year}{2020}\natexlab{}.
\newblock \showarticletitle{Adversarial attacks against network intrusion
  detection in IoT systems}.
\newblock \bibinfo{journal}{\emph{IEEE Internet of Things Journal}}
  \bibinfo{volume}{8}, \bibinfo{number}{13} (\bibinfo{year}{2020}),
  \bibinfo{pages}{10327--10335}.
\newblock


\bibitem[Rozemberczki and Sarkar(2020)]%
        {10.1145/3340531.3411866}
\bibfield{author}{\bibinfo{person}{Benedek Rozemberczki} {and}
  \bibinfo{person}{Rik Sarkar}.} \bibinfo{year}{2020}\natexlab{}.
\newblock \showarticletitle{Characteristic Functions on Graphs: Birds of a
  Feather, from Statistical Descriptors to Parametric Models}. In
  \bibinfo{booktitle}{\emph{Proceedings of the 29th ACM International
  Conference on Information and Knowledge Management}} (Virtual Event, Ireland)
  \emph{(\bibinfo{series}{CIKM '20})}. \bibinfo{publisher}{Association for
  Computing Machinery}, \bibinfo{address}{New York, NY, USA},
  \bibinfo{pages}{1325–1334}.
\newblock
\showISBNx{9781450368599}
\urldef\tempurl%
\url{https://doi.org/10.1145/3340531.3411866}
\showDOI{\tempurl}


\bibitem[Satam and Hariri(2020)]%
        {satam2020wids}
\bibfield{author}{\bibinfo{person}{Pratik Satam} {and} \bibinfo{person}{Salim
  Hariri}.} \bibinfo{year}{2020}\natexlab{}.
\newblock \showarticletitle{WIDS: An anomaly based intrusion detection system
  for Wi-Fi (IEEE 802.11) protocol}.
\newblock \bibinfo{journal}{\emph{IEEE Transactions on Network and Service
  Management}} \bibinfo{volume}{18}, \bibinfo{number}{1}
  (\bibinfo{year}{2020}), \bibinfo{pages}{1077--1091}.
\newblock


\bibitem[Sharafaldin et~al\mbox{.}(2018)]%
        {sharafaldin2018toward}
\bibfield{author}{\bibinfo{person}{Iman Sharafaldin},
  \bibinfo{person}{Arash~Habibi Lashkari}, {and} \bibinfo{person}{Ali~A
  Ghorbani}.} \bibinfo{year}{2018}\natexlab{}.
\newblock \showarticletitle{Toward generating a new intrusion detection dataset
  and intrusion traffic characterization.}
\newblock \bibinfo{journal}{\emph{ICISSp}}  \bibinfo{volume}{1}
  (\bibinfo{year}{2018}), \bibinfo{pages}{108--116}.
\newblock


\bibitem[Stiawan et~al\mbox{.}(2020)]%
        {stiawan2020cicids}
\bibfield{author}{\bibinfo{person}{Deris Stiawan}, \bibinfo{person}{Mohd
  Yazid~Bin Idris}, \bibinfo{person}{Alwi~M Bamhdi}, \bibinfo{person}{Rahmat
  Budiarto}, {et~al\mbox{.}}} \bibinfo{year}{2020}\natexlab{}.
\newblock \showarticletitle{CICIDS-2017 dataset feature analysis with
  information gain for anomaly detection}.
\newblock \bibinfo{journal}{\emph{IEEE Access}}  \bibinfo{volume}{8}
  (\bibinfo{year}{2020}), \bibinfo{pages}{132911--132921}.
\newblock


\bibitem[Tavallaee et~al\mbox{.}(2009)]%
        {tavallaee2009detailed}
\bibfield{author}{\bibinfo{person}{Mahbod Tavallaee}, \bibinfo{person}{Ebrahim
  Bagheri}, \bibinfo{person}{Wei Lu}, {and} \bibinfo{person}{Ali~A Ghorbani}.}
  \bibinfo{year}{2009}\natexlab{}.
\newblock \showarticletitle{A detailed analysis of the KDD CUP 99 data set}. In
  \bibinfo{booktitle}{\emph{2009 IEEE symposium on computational intelligence
  for security and defense applications}}. Ieee, \bibinfo{pages}{1--6}.
\newblock


\bibitem[Tavallaee et~al\mbox{.}(2010)]%
        {tavallaee2010toward}
\bibfield{author}{\bibinfo{person}{Mahbod Tavallaee}, \bibinfo{person}{Natalia
  Stakhanova}, {and} \bibinfo{person}{Ali~Akbar Ghorbani}.}
  \bibinfo{year}{2010}\natexlab{}.
\newblock \showarticletitle{Toward credible evaluation of anomaly-based
  intrusion-detection methods}.
\newblock \bibinfo{journal}{\emph{IEEE Transactions on Systems, Man, and
  Cybernetics, Part C (Applications and Reviews)}} \bibinfo{volume}{40},
  \bibinfo{number}{5} (\bibinfo{year}{2010}), \bibinfo{pages}{516--524}.
\newblock


\bibitem[Viriyasitavat et~al\mbox{.}(2019)]%
        {8764459}
\bibfield{author}{\bibinfo{person}{Wattana Viriyasitavat}, \bibinfo{person}{Li
  Da~Xu}, \bibinfo{person}{Zhuming Bi}, {and} \bibinfo{person}{Assadaporn
  Sapsomboon}.} \bibinfo{year}{2019}\natexlab{}.
\newblock \showarticletitle{New Blockchain-Based Architecture for Service
  Interoperations in Internet of Things}.
\newblock \bibinfo{journal}{\emph{IEEE Transactions on Computational Social
  Systems}} \bibinfo{volume}{6}, \bibinfo{number}{4} (\bibinfo{year}{2019}),
  \bibinfo{pages}{739--748}.
\newblock
\urldef\tempurl%
\url{https://doi.org/10.1109/TCSS.2019.2924442}
\showDOI{\tempurl}


\bibitem[Wu et~al\mbox{.}(2023a)]%
        {10026337}
\bibfield{author}{\bibinfo{person}{Jiashu Wu}, \bibinfo{person}{Hao Dai},
  \bibinfo{person}{Yang Wang}, \bibinfo{person}{Kejiang Ye}, {and}
  \bibinfo{person}{Chengzhong Xu}.} \bibinfo{year}{2023}\natexlab{a}.
\newblock \showarticletitle{Heterogeneous Domain Adaptation for IoT Intrusion
  Detection: A Geometric Graph Alignment Approach}.
\newblock \bibinfo{journal}{\emph{IEEE Internet of Things Journal}}
  (\bibinfo{year}{2023}), \bibinfo{pages}{1--1}.
\newblock
\urldef\tempurl%
\url{https://doi.org/10.1109/JIOT.2023.3239872}
\showDOI{\tempurl}


\bibitem[Wu et~al\mbox{.}(2022)]%
        {9833301}
\bibfield{author}{\bibinfo{person}{Jiashu Wu}, \bibinfo{person}{Hao Dai},
  \bibinfo{person}{Yang Wang}, \bibinfo{person}{Yong Zhang},
  \bibinfo{person}{Dong Huang}, {and} \bibinfo{person}{Chengzhong Xu}.}
  \bibinfo{year}{2022}\natexlab{}.
\newblock \showarticletitle{PackCache: An Online Cost-Driven Data Caching
  Algorithm in the Cloud}.
\newblock \bibinfo{journal}{\emph{IEEE Trans. Comput.}} (\bibinfo{year}{2022}),
  \bibinfo{pages}{1--8}.
\newblock
\showISSN{1557-9956}
\urldef\tempurl%
\url{https://doi.org/10.1109/TC.2022.3191969}
\showDOI{\tempurl}


\bibitem[Wu et~al\mbox{.}(2023b)]%
        {10082914}
\bibfield{author}{\bibinfo{person}{Jiashu Wu}, \bibinfo{person}{Yang Wang},
  \bibinfo{person}{Hao Dai}, \bibinfo{person}{Chengzhong Xu}, {and}
  \bibinfo{person}{Kenneth~B. Kent}.} \bibinfo{year}{2023}\natexlab{b}.
\newblock \showarticletitle{Adaptive Bi-Recommendation and Self-Improving
  Network for Heterogeneous Domain Adaptation-Assisted IoT Intrusion
  Detection}.
\newblock \bibinfo{journal}{\emph{IEEE Internet of Things Journal}}
  (\bibinfo{year}{2023}), \bibinfo{pages}{1--1}.
\newblock
\urldef\tempurl%
\url{https://doi.org/10.1109/JIOT.2023.3262458}
\showDOI{\tempurl}


\bibitem[Wu et~al\mbox{.}(2023c)]%
        {9933783}
\bibfield{author}{\bibinfo{person}{Jiashu Wu}, \bibinfo{person}{Yang Wang},
  \bibinfo{person}{Binhui Xie}, \bibinfo{person}{Shuang Li},
  \bibinfo{person}{Hao Dai}, \bibinfo{person}{Kejiang Ye}, {and}
  \bibinfo{person}{Chengzhong Xu}.} \bibinfo{year}{2023}\natexlab{c}.
\newblock \showarticletitle{Joint Semantic Transfer Network for IoT Intrusion
  Detection}.
\newblock \bibinfo{journal}{\emph{IEEE Internet of Things Journal}}
  \bibinfo{volume}{10}, \bibinfo{number}{4} (\bibinfo{year}{2023}),
  \bibinfo{pages}{3368--3383}.
\newblock
\urldef\tempurl%
\url{https://doi.org/10.1109/JIOT.2022.3218339}
\showDOI{\tempurl}


\bibitem[Xie et~al\mbox{.}(2022)]%
        {xie2022collaborative}
\bibfield{author}{\bibinfo{person}{Binhui Xie}, \bibinfo{person}{Shuang Li},
  \bibinfo{person}{Fangrui Lv}, \bibinfo{person}{Chi~Harold Liu},
  \bibinfo{person}{Guoren Wang}, {and} \bibinfo{person}{Dapeng Wu}.}
  \bibinfo{year}{2022}\natexlab{}.
\newblock \showarticletitle{A collaborative alignment framework of transferable
  knowledge extraction for unsupervised domain adaptation}.
\newblock \bibinfo{journal}{\emph{IEEE Transactions on Knowledge and Data
  Engineering}} (\bibinfo{year}{2022}).
\newblock


\bibitem[Yao et~al\mbox{.}(2019)]%
        {yao2019capsule}
\bibfield{author}{\bibinfo{person}{Haipeng Yao}, \bibinfo{person}{Pengcheng
  Gao}, \bibinfo{person}{Jingjing Wang}, \bibinfo{person}{Peiying Zhang},
  \bibinfo{person}{Chunxiao Jiang}, {and} \bibinfo{person}{Zhu Han}.}
  \bibinfo{year}{2019}\natexlab{}.
\newblock \showarticletitle{Capsule network assisted IoT traffic classification
  mechanism for smart cities}.
\newblock \bibinfo{journal}{\emph{IEEE Internet of Things Journal}}
  \bibinfo{volume}{6}, \bibinfo{number}{5} (\bibinfo{year}{2019}),
  \bibinfo{pages}{7515--7525}.
\newblock


\end{thebibliography}

\newpage
\appendix

\section{Appendix}\label{sec:appendix}

\subsection{\rw{Acronym Table}}\label{sec:acronym_table}

\begin{table}[!ht]
    \caption{\rw{The acronym table and the corresponding interpretation (Based on the order of appearance in the paper. )}}
    \vspace{-3mm}
    \centering
    \label{tab:acronym_table}
    \rw{\begin{tabular}{c|l}
    \Xhline{2\arrayrulewidth}
    Acronym & Interpretation \\ \hline
    OSDN & Open-Set Dandelion Network \\
    DA & Domain Adaptation \\
    NI & Network Intrusion \\
    II & IoT Intrusion \\
    OSDA & Open-Set Domain Adaptation \\
    DASM & Dandelion Angular Separation Mechanism \\
    DEAM & Dandelion Embedding Alignment Mechanism \\
    DSDM & Discriminating Sampled Dandelion Mechanism \\
    SDCM & Semantic Dandelion Correction Mechanism \\
    ML & Machine Learning \\
    DL & Deep Learning \\
    CS & Cosine Similarity \\
    EA & Embedding Alignment \\
    CP & Compactness \\
    SUP & Supervision \\
    SM & Semantic \\
    SC & Semantic Correction \\
    CE & Cross Entropy \\
    \Xhline{2\arrayrulewidth}
    \end{tabular}}
\end{table}

\subsection{Notation Table}\label{notation_table}

\begin{table}[!ht]
  \caption{\rw{The notation table and the corresponding interpretation (Based on the order of appearance in the paper. )}}
  \vspace{-3mm}
  \centering
  \label{tab:notation_table}
  \rw{\begin{tabular}{c|l}
    \Xhline{2\arrayrulewidth}
    Notation & Interpretation \\ \hline
    $\mathcal{D}_{S}$ & Source NI domain \\
    $\mathcal{X}_{S}$ & Source NI domain traffic features \\
    $\mathcal{Y}_{S}$ & Source NI domain traffic intrusion labels \\
    $x_{S_i}$ & The $i$\textsuperscript{th} traffic instance in $\mathcal{X}_{S}$ \\
    $y_{S_i}$ & The intrusion label of $x_{S_i}$ \\
    $n_S$ & Number of instances in $\mathcal{X}_{S}$ \\
    $d_S$ & Instance dimension of $\mathcal{X}_{S}$ \\
    $K$ & Number of intrusion categories in $\mathcal{D}_{S}$ \\
    $K'$ & Number of intrusion categories in $\mathcal{D}_{T}$ \\
    $f(x_i)$ & The feature projector \\
    $E_{S}$ & The source feature projector \\
    $E_{T}$ & The target feature projector \\
    $d_{C}$ & The dimension of the common feature subspace \\
    $d_{max}^{(i)}$ & The maximum intra-category deviation of source intrusion category $i$ \\
    $COS()$ & Cosine Similarity \\
    $n_S^{(i)}$ & Number of instances in the $i$\textsuperscript{th} source intrusion category \\
    $\mu_{S}^{(i)}$ & Mean of the source intrusion category $i$ \\
    $x_{S_j}^{(i)}$ & The $j$\textsuperscript{th} instance of source $i$\textsuperscript{th} intrusion category \\
    $y_{T_j}^{D}$ & The dandelion-based membership for the $j$\textsuperscript{th} target instance $x_{T_j}$ \\
    $CS_{S}$ & The source category pair-wise Cosine similarity matrix \\
    $CS_{S}^{ij}$ & The Cosine similarity between the $i$\textsuperscript{th} and $j$\textsuperscript{th} source intrusion category \\
    $\mathcal{L}_{SS}$ & Source dandelion separation loss \\
    $\mathcal{L}_{ST}$ & Target dandelion separation loss \\
    $G_{S}$ & The source dandelion graph \\
    $V_{S}$ & Vertices in $G_{S}$ \\
    $E_{G}$ & Edges in $G_{S}$ \\
    $V_{S}^{(i)}$ & The $i$\textsuperscript{th} vertex in the $G_{S}$ \\
    $E_{S}^{ij}$ & The edge connecting $V_{S}^{(i)}$ and $V_{S}^{(j)}$ \\
    $\vmathbb{0}$ & The origin \\
    $\mathcal{L}_{EA}$ & Dandelion embedding alignment loss \\
    $\phi_{S}$ & The graph embedding of the source domain dandelion \\
    $\mathcal{L}_{CP}$ & Discriminating sampled dandelion loss \\
    $D()$ & The discriminator \\
    $G_{\mathcal{DD}_{S}}$ & The graph embedding of the source dandelion \\
    $G_{\mathcal{DD}_{T}}$ & The graph embedding of the target dandelion \\
    $G_{\mathcal{DD}_{*}^{j}}$ & The graph embedding of the $j$\textsuperscript{th} sampled dandelion \\
    $N$ & The amount of child dandelion being sampled \\
    $\mathcal{L}_{SUP}$ & The overall supervision loss \\
    $\mathcal{L}_{SUP_S}$ & The source supervision loss \\
    $\mathcal{L}_{SUP_U}$ & The unknown supervision loss \\
    $\mathcal{L}_{CE}$ & The cross entropy loss \\
    $n_R$ & The amount of unknown instances being generated \\
    $\mathcal{X}_{R}$ & The generated unknown instances for unknown training \\
    $C$ & The intrusion classifier \\
    \Xhline{2\arrayrulewidth}
  \end{tabular}}
\end{table}

\begin{table}[!ht]
  \caption{\rw{The notation table and the corresponding interpretation (Continued)}}
  \vspace{-3mm}
  \centering
  \label{tab:notation_table_continued}
  \rw{\begin{tabular}{c|l}
  \Xhline{2\arrayrulewidth}
  Notation & Interpretation \\ \hline
  $p_{S_j}^{(i)}$ & The probabilistic semantic of the $j$\textsuperscript{th} source instance in category $i$ \\
  $\mathcal{DD}_{\mathcal{S}S}$ & The source semantic dandelion \\
  $\mathcal{DD}_{\mathcal{S}S}^{(i)}$ & The $i$\textsuperscript{th} pappus of the source semantic dandelion \\
  $CS_{SM}$ & The Cosine similarity matrix between semantic dandelions \\
  $CS_{SM}^{ij}$ & The Cosine similarity between $\mathcal{DD}_{\mathcal{S}S}^{(i)}$ and $\mathcal{DD}_{\mathcal{S}T}^{(j)}$ \\
  $\mathcal{L}_{SC}$ & The semantic dandelion correction loss \\
  $\alpha_{S}, \alpha_{U}$ & Hyperparameter controlling $\mathcal{L}_{SUP_S}$ and $\mathcal{L}_{SUP_U}$, respectively \\
  $\beta_{S}$, $\beta_{T}$ & Hyperparameter controlling $\mathcal{L}_{SS}$ and $\mathcal{L}_{ST}$, respectively \\
  $\delta$ & Hyperparameter controlling $\mathcal{L}_{EA}$ \\
  $\theta$ & Hyperparameter controlling $\mathcal{L}_{SC}$ \\
  $\gamma$ & Hyperparameter controlling $\mathcal{L}_{CP}$ \\
  $TP^{(k)}$ & True positive of category $k$ \\
  $|\mathcal{X}_T^{(k)}|$ & Number of target instances in intrusion category $k$ \\
  $\mathcal{O}$ & The openness level \\
  $\mathcal{DD}_{S\cup T}$ & The source-target combined dandelion \\
  $CS_{S\cup T}$ & The inter-pappus Cosine similarity matrix of $\mathcal{DD}_{S\cup T}$ \\
  $\mu_{S\cup T}^{(i)}$ & The $i$\textsuperscript{th} pappus of $CS_{S\cup T}$ \\
  $CS_{S\cup T}^{ij}$ & The Cosine similarity between $\mu_{S\cup T}^{(i)}$ and $\mu_{S\cup T}^{(j)}$ \\
  $d_{max}$ & The average category-wise maximum deviation \\
  \Xhline{2\arrayrulewidth}
  \end{tabular}}
\end{table}

\end{document}